\def\paperID{}
\def\confName{CVPR}
\def\confYear{2026}

\def\paperTitle{Tango: Taming Visual Signals for Efficient Video Large Language Models}

\def\authorBlock{
   Shukang Yin$^{1}$, ~Sirui Zhao$^{1\ast}$, ~Hanchao Wang$^{2}$, ~Baozhi Jia$^{2}$, 
   ~Xianquan Wang$^{1}$,
   \\Chaoyou Fu$^{3}$, ~Enhong Chen$^{1}$\thanks{Corresponding authors.} \\[0.5em]
$^1$University of Science and Technology of China, $^2$Reconova AI Lab, $^3$Nanjing University \\
Code: \href{https://github.com/xjtupanda/Tango}{github.com/xjtupanda/Tango}
}

\newif\ifarxiv \newcommand{\arxiv}{\arxivtrue}

\arxiv

\pdfoutput=1
\PassOptionsToPackage{svgnames}{xcolor}
\documentclass[10pt,twocolumn,letterpaper]{article}
\ifarxiv \usepackage[pagenumbers]{cvpr} \fi

\usepackage[utf8]{inputenc}
\usepackage{tikz}
\usetikzlibrary{arrows.meta, calc, decorations.pathreplacing}
\usepackage{graphicx}
\usepackage[numbers,sort&compress]{natbib}
\usepackage{indentfirst}   % fit format. indent the first paragraph of the section
\usepackage{bbding, pifont}
\usepackage{bm}
\usepackage{siunitx}
\usepackage{mathtools}
\usepackage{amsmath}	
\usepackage{amssymb}	
\usepackage{booktabs}
\usepackage{times}
\usepackage{microtype}
\usepackage{epsfig}
\usepackage{xcolor}
\usepackage{caption}
\usepackage{transparent}
\usepackage{float}
\usepackage{placeins}
\usepackage{color, colortbl}
\usepackage{stfloats}
\usepackage{enumitem}
\usepackage{tabularx}

\usepackage[normalem]{ulem}
\useunder{\uline}{\ul}{}

\usepackage{xstring}
\usepackage{multirow}
\usepackage{makecell}
\usepackage{xspace}
\usepackage{url}
\usepackage{subcaption}
\usepackage{tcolorbox}
\usepackage[hang,flushmargin]{footmisc}

\def\eg{\emph{e.g}\onedot} 
\def\ie{\emph{i.e}\onedot} 
 
 \def\vs{\emph{vs}\onedot}
\def\wrt{w.r.t\onedot}

\makeatletter\renewcommand\paragraph{\@startsection{paragraph}{4}{\z@}{.5em\@plus1ex\@minus.2ex}{-.5em}{\normalfont\normalsize\bfseries}}
\ifarxiv  \fi

\newcommand{\R}[1]{{%
    \textbf{%
        \ifstrequal{#1}{1}{\textcolor{red}{R#1}}{%
        \ifstrequal{#1}{2}{\textcolor{blue}{R#1}}{%
        \ifstrequal{#1}{3}{\textcolor{magenta}{R#1}}{%
        \ifstrequal{#1}{4}{\textcolor{teal}{R#1}}{%
                           \textcolor{cyan}{R#1}%
        }}}}%
    }%
}}

\usepackage{amsmath,amssymb,mathtools}
\usepackage{amsthm}

\usepackage[table]{xcolor}

\definecolor{main_block}{HTML}{FBE3D6}
\definecolor{attn_color}{HTML}{87AE73}
\definecolor{p-attn_color}{HTML}{CA7B80}

\newcommand{\tablestyle}[2]{\setlength{\tabcolsep}{#1}\renewcommand{\arraystretch}{#2}}
\newcommand{\shline}{\specialrule{.1em}{.05em}{.05em}}

\colorlet{grey}{gray}         % alias so \textcolor{grey}{...} works
\colorlet{MidnightBlue}{MidnightBlue}
\definecolor{isabelline}{HTML}{EAEAEA} % light neutral gray banding
\definecolor{lightblue}{HTML}{E6F1FF}  % gentle light blue row highlight

\newcommand{\cmark}{\textcolor{OliveGreen}{\ding{51}}}%
\newcommand{\xmark}{\textcolor{Maroon}{\ding{55}}}%

\usepackage{algorithm}
\usepackage{algpseudocode}
\usepackage{ragged2e}
\algrenewcommand\algorithmicrequire{\textbf{Input:}}
\algrenewcommand\algorithmicensure{\textbf{Output:}}
\algrenewcommand{\algorithmiccomment}[1]{\hfill $\triangleright$ \small #1}
\usepackage{xr-hyper}

\makeatletter
\newcommand*{\addFileDependency}[1]{
  \typeout{(#1)}
  \@addtofilelist{#1}
  \IfFileExists{#1}{}{\typeout{No file #1.}}
}

\makeatother

\definecolor{cvprblue}{rgb}{0.21,0.49,0.74}
\usepackage[pagebackref,breaklinks,colorlinks,allcolors=cvprblue]{hyperref}
\usepackage[capitalize]{cleveref}
\crefname{section}{Sec.}{Secs.}
\crefname{table}{Table}{Tables}
\crefname{figure}{Fig.}{Figs.}

\ifarxiv \crefname{appendix}{App.}{Apps.}
\else \crefname{appendix}{Suppl.}{Suppls.} \fi

\begin{document}
%% TITLE
\title{\paperTitle}
\author{\authorBlock}
\maketitle

\begin{abstract}
Token pruning has emerged as a mainstream approach for developing efficient Video Large Language Models (Video LLMs). 
This work revisits and advances the two predominant token-pruning paradigms: attention-based selection and similarity-based clustering. Our study reveals two critical limitations in existing methods: (1) conventional top-k selection strategies fail to fully account for the attention distribution, which is often spatially multi-modal and long-tailed in magnitude; and (2) direct similarity-based clustering frequently generates fragmented clusters, resulting in distorted representations after pooling. 
To address these bottlenecks, we propose \textbf{Tango}, a novel framework designed to optimize the utilization of visual signals. 
Tango integrates a diversity-driven strategy to enhance attention-based token selection, and introduces Spatio-temporal Rotary Position Embedding (ST-RoPE) to preserve geometric structure via locality priors.
Comprehensive experiments across various Video LLMs and video understanding benchmarks demonstrate the effectiveness and generalizability of our approach. Notably, when retaining only 10\% of the video tokens, Tango preserves 98.9\% of the original performance on LLaVA-OV while delivering a 1.88$\times$ inference speedup.
\end{abstract}

\section{Introduction}
\label{sec:intro}

\begin{figure}[!th]
    \centering
    \setlength{\abovecaptionskip}{2mm}
    \setlength{\belowcaptionskip}{-3mm}
    \includegraphics[width=\columnwidth]{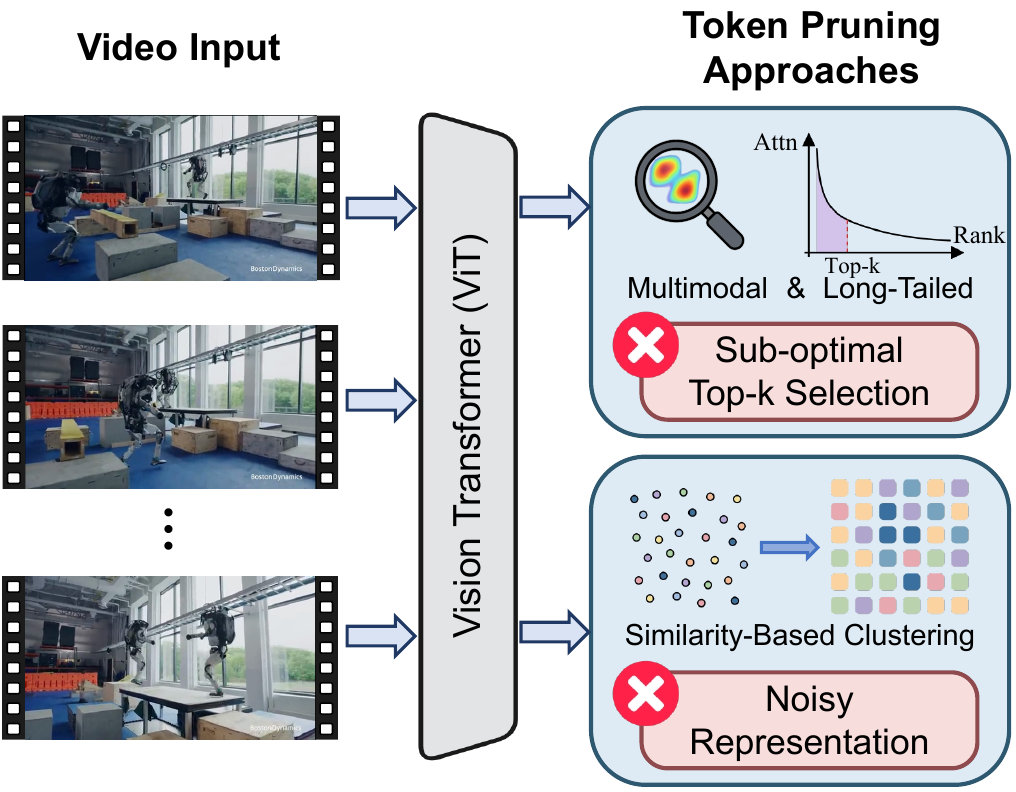} 
    \caption{\textbf{Limitations of two typical pre-LLM token pruning approaches.}
    (Top Right) Top-k selection fails to fully capture the attention distribution, which is spatially multimodal and long-tailed in magnitude.
    (Bottom Right) Direct similarity-based clustering can result in noisy representations.}
    \label{fig:teaser}
\end{figure}

Recent years have witnessed remarkable advancements in efficient Video Large Language Models (Video LLMs) through training-free token pruning techniques~\cite{fastv, dart, visionzip, vidcom2, fastvid}.
Within this domain, previous work primarily focuses on two pivotal objectives for token pruning: \textit{saliency}~\cite{fastv, pyramiddrop, visionzip} and \textit{diversity}~\cite{dart, vidcom2, divprune}.
Specifically, saliency-based approaches typically rely on \textit{attention weights} to identify critical tokens for retention, while diversity-based methods aim to minimize feature redundancy within the retained set, typically measured via \textit{token similarity}.

In this work, we conduct an in-depth investigation into these two visual signals, grounded in fundamental token-pruning paradigms.
As illustrated in~\cref{fig:teaser}, our analysis reveals two critical limitations in existing approaches:
(1) The distribution of attention scores is inherently multi-modal and long-tailed, making a rigid Top-$k$ selection strategy sub-optimal for capturing this pattern.
(2) Direct similarity-based clustering on tokens yields noisy object representations.

To address these shortcomings, we propose targeted improvements: 
First, to enhance mode and tail coverage during salient token selection, we introduce a diversity-driven selection process. This involves clustering an expanded set of candidate tokens, followed by an intra-cluster selection step to ensure comprehensive semantic coverage.
Second, to better preserve the geometric structure of visual objects, we inject a locality prior into the inter-token similarity calculation, which explicitly penalizes similarities between spatio-temporally distant tokens.
Concretely, we design a Spatio-temporal Rotary Position Embedding (ST-RoPE) mechanism that encodes relative distances along both the spatial and temporal axes. 
We demonstrate that this formulation can be seamlessly integrated into standard clustering frameworks. 

Comprehensive experiments validate the effectiveness of our approach. Notably, retaining only 10\% video tokens preserves 98.9\% of the full performance of LLaVA-OneVision-7B while accelerating inference speed by 1.88$\times$.

In summary, our main contributions are threefold:
\begin{itemize}
    \item We identify and analyze critical limitations in how current video token pruning methods utilize attention weights and inter-token similarity.
    \item We propose a novel pruning framework featuring a diversity-driven salient token selection strategy and ST-RoPE, a specialized position-encoding mechanism injecting spatio-temporal awareness into similarity calculations.
    \item We conduct extensive experiments demonstrating the effectiveness of our method and offer valuable insights for future efficient Video LLM design.
\end{itemize}

\section{Related Work}
\label{sec:related}
Vision token pruning methods developed in recent years can be organized along two primary dimensions: pruning stages and pruning criteria.
In this section, we discuss and compare the key techniques of representative works.

\subsection{Token Pruning Stages}
Depending on where vision tokens are pruned, existing methods can be categorized into pre-LLM methods and intra-LLM methods, in which video tokens are pruned before LLM processing or within the LLM backbone.

Pre-LLM methods prune vision tokens after the vision encoder but before the LLM backbone. 
These prompt-agnostic methods use visual signals, such as similarities, for pruning. 
For instance, PruneVid~\cite{prunevid} groups consecutive frames into segments and exploits cross-frame similarities to compress static tokens, better preserving dynamic signals. 
HoliTom~\cite{holitom} advances this idea by formulating video segmentation as a static-token maximization problem and solving it via dynamic programming. 
In contrast, intra-LLM methods prune within the LLM backbone, capturing cross-modal interactions between textual prompts and vision tokens, utilizing text-to-image attention scores~\cite{fastv} or token similarities~\cite{dart}.

\subsection{Token Pruning Criteria}
Existing works primarily adopt two types of token selection criterion, \ie, saliency and diversity. We organize representative works according to the main criterion.

\paragraph{Saliency-Based Methods.} 
This line of work primarily relies on specific metrics, typically attention scores, to identify and retain critical tokens.
FastV~\cite{fastv} computes attention weights from prompt to vision tokens within a specific LLM layer, retaining those with the highest scores.
PyramidDrop~\cite{pyramiddrop} refines this approach via a multi-stage strategy that progressively increases pruning ratios across deeper LLM layers.

\paragraph{Diversity-Based Methods.}
In contrast to saliency-based works, diversity-based methods aim to minimize information redundancy among the retained vision tokens. Cosine similarity acts as the predominant metric for quantifying token resemblance. 
For instance, DART~\cite{dart} initializes a set of pivot tokens and iteratively incorporates new tokens that exhibit the highest dissimilarity to the current selection. 
Similarly, VidCom\textsuperscript{2}~\cite{vidcom2} employs a two-stage pruning framework that explicitly targets frame-wise and token-wise uniqueness in the first and second stages, respectively.

\paragraph{Hybrid Methods.}
Recent endeavors have explored hybrid approaches that integrate the strengths of saliency-based selection and diversity-aware pruning~\cite{fu2025framefusion, visionzip}. 
For example, VisionZip~\cite{visionzip} retains important tokens based on attention scores and merges contextually similar tokens. 
FastVID~\cite{fastvid} and HoliTom~\cite{holitom} improve this framework in the video domain, employing clustering algorithms to effectively aggregate spatio-temporally redundant tokens. 
Our approach advances this paradigm by introducing a more effective salient token selection strategy and a mechanism that better preserves geometric structure in the clustering process.
\begin{figure*}[!thb]\centering	\includegraphics[width=0.96\textwidth]{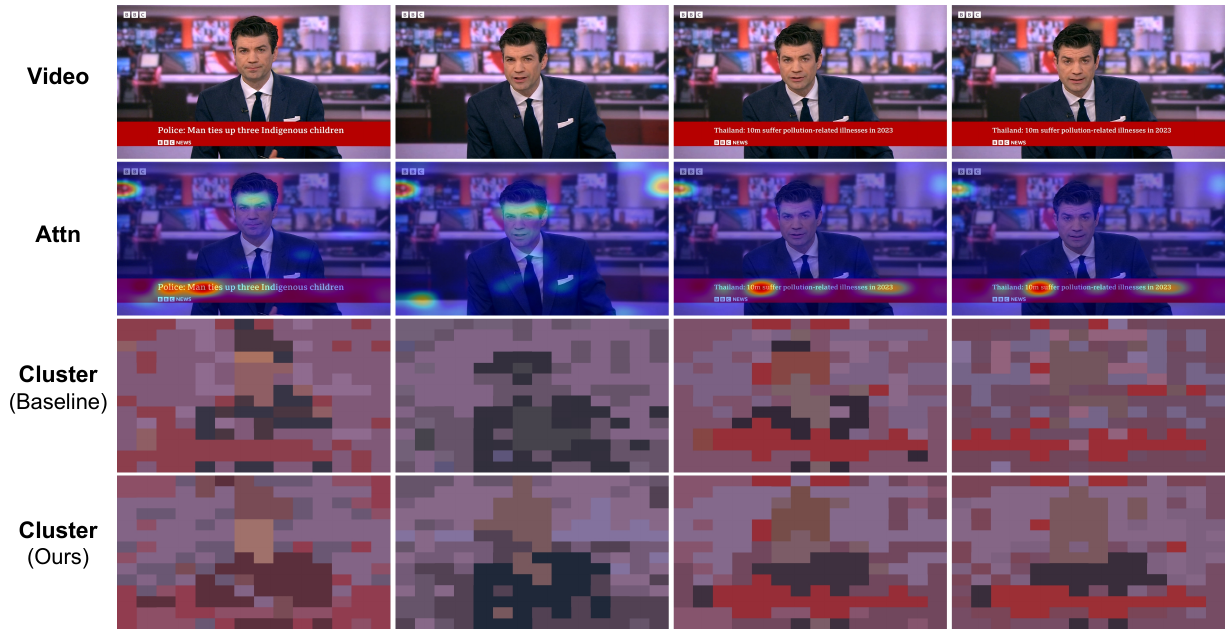}
 \caption{\textbf{Illustration of attention distribution and similarity-based clustering.}
    \textit{Attention Heatmaps:} The spatial distribution of attention scores exhibits multiple modes (\ie, local maxima of attention scores) corresponding to distinct semantic regions (\eg, subtitle and head in the first frame). 
    \textit{Clustering Results:} Direct similarity-based cluster assignment (Baseline) leads to spatially fragmented results, where a coherent semantic entity (\eg, the head or body) is split into different clusters or mixed with background. Pooling on these clusters may result in noisy representations. 
    In comparison, our method (bottom row) mitigates this issue and exhibits superior geometric shape preservation. 
    Visualization is extracted using the SigLIP~\cite{siglip} ViT of LLaVA-OV-7B on a Video-MME sample (\textcolor{gray}{vid: lmlw4NNZBbw}). More cases and details are provided in the Appendix.
}
	\label{fig:motivation}
\end{figure*}

\section{Preliminaries}
\paragraph{Architecture of Video LLMs.}
We adhere to the prevailing paradigm for large multimodal models~\cite{yin2024survey}, which comprises a visual encoder, a projector, and an LLM backbone.

Given an input video $\mathcal{V}$ consisting of $T$ frames, denoted as $\mathcal{V} = \{v_t\}_{t=1}^T$, a pre-trained vision encoder (\eg, ViT) extracts visual representations $\mathbf{H}^v \in \mathbb{R}^{T \times H \times W \times D}$, where $N_v$ is the number of tokens per frame and $D$ is the hidden dimension of the encoder.
These vision features are then projected (and possibly pooled), resulting in $\mathbf{X}^v \in \mathbb{R}^{T \times H^\prime \times W^\prime \times d}$, with $N_{v}^{\prime}=H^\prime \times W^\prime \le N_v=H \times W$, where $d$ is the hidden dimension of the LLM.

These visual tokens are concatenated with the text embeddings of a user instruction, denoted as $\mathbf{X}^q \in \mathbb{R}^{L_t \times d}$, forming a unified multimodal input sequence $\mathbf{X} = [\mathbf{X}^v; \mathbf{X}^q]$. Given this input, the LLM backbone then outputs a response $\mathbf{Y} = \{\boldsymbol{y}_j\}_{j=1}^L$ autoregressively:
\begin{equation*}
p(\mathbf{Y} \mid \mathbf{X}) = \prod_{j=1}^{L} p(\boldsymbol{y}_j \mid \mathbf{X}, \mathbf{Y}_{<j})
\text{.}
\end{equation*}

\paragraph{Computational Complexity.}
The main computational bottleneck lies at the LLM backbone, which is a stack of Transformer Layers. When only examining the main computation components, \ie, self-attention and feed-forward network (FFN), the total floating-point operations (FLOPs) are:
% can be expressed as:
\begin{align*}
\text{FLOPs}(\texttt{Layer}) &= \underbrace{4Nd^2\left(1+\frac{G}{H}\right) + 4N^2d}_{\text{Self-Attention (GQA)}} \notag \\
&\quad + \underbrace{6Ndm}_{\text{FFN (SwiGLU)}} \notag \\
&= N \cdot \left[4d^2\left(1+\frac{G}{H}\right) + 6dm\right] + N^2(4d) \notag \\
\text{FLOPs}(\texttt{LLM}) &= L \times \text{FLOPs}(\texttt{Layer})\text{,}
\label{eq:flops}
\end{align*}
where $L$ denotes the number of Transformer layers, $N$ the sequence length, $d$ the hidden size, and $m$ the intermediate size of the feed-forward layer. Additionally, $H$ and $G$ denote the number of query heads and key-value groups, respectively.

To ensure a fair comparison of pruning performance, we align retention ratios based on similar LLM FLOPs. 
Since the coefficient of the linear term significantly outweighs the quadratic term (\eg, $\sim 32,500 \times$ in the Qwen2-7B model~\cite{qwen2}), in common settings (context shorter than 8K), we can simplify the \textit{equivalent retention ratio}~\cite{sparsevlm} to the average of retention ratios across all LLM layers:
\begin{equation}
\begin{split}
% \bar{r}= \dfrac{l_p + (L-l_p)\times r}{L}
\bar{r} = \dfrac{\sum_{l=1}^{L} r_l}{L}
\text{,}
\end{split}
\label{eq:ret_ratio}
\end{equation}
where $r_l$ is the \textit{actual retention ratio} for the $l$-th layer.

\begin{figure}[!t]
    \centering
    \setlength{\abovecaptionskip}{1mm}
    \setlength{\belowcaptionskip}{-4mm}
    \includegraphics[width=0.92\columnwidth]{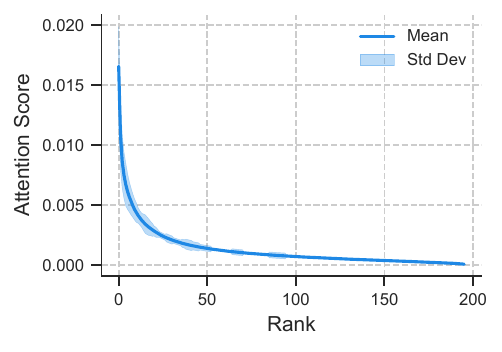} 
    \caption{\textbf{Distribution of sorted attention scores.}
    The scores exhibit a distinct long-tailed distribution with a narrow variance band, demonstrating highly stable attention patterns across videos. Results are averaged over 100 video samples from ActivityNet~\cite{caba2015activitynet}.}
    \label{fig:attn_dist}
\end{figure}

\paragraph{Investigating Visual Signals.}
We investigate the characteristics of visual signals grounded in mainstream video token pruning paradigms, namely attention-based salient token selection and similarity-based clustering and merging.

Analyzing the attention distribution, we observe that the heatmaps exhibit multi-modal patterns (\ie, concentrated high-attention regions, as seen in the top two rows of~\cref{fig:motivation}) that naturally align with distinct semantic entities, such as the subtitle and the newscaster's head. Moreover, the attention scores follow a long-tailed distribution (\cref{fig:attn_dist}), with top-ranking tokens consistently accounting for the majority of the attention weights. 
It is worth noting that some of these high-ranking tokens are attention sinks~\cite{darcet2023vision,jiang2025vision} that consistently hold the highest attention scores and remain in fixed spatial positions (\#28 and \#26 are Top-1 in LLaVA-OV-7B and LLaVA-Video-7B, respectively). Visually, these sink tokens manifest as the persistent high-activation region in the corner (see the upper-left corner of frames in the second row of~\cref{fig:motivation}). We provide further discussion on this phenomenon in the Appendix.

In the clustering results, we observe that direct similarity-based clustering leads to spatially fragmented clusters (See comparisons in the last two rows of~\cref{fig:motivation}). Pooling on fragmented clusters may result in noisy representations, losing the shape structure of objects.
These observations motivate our proposed method.

\begin{figure}[!t]
\centering
 \setlength{\abovecaptionskip}{3mm}
    \setlength{\belowcaptionskip}{-2mm}
\includegraphics[width=0.98\columnwidth]{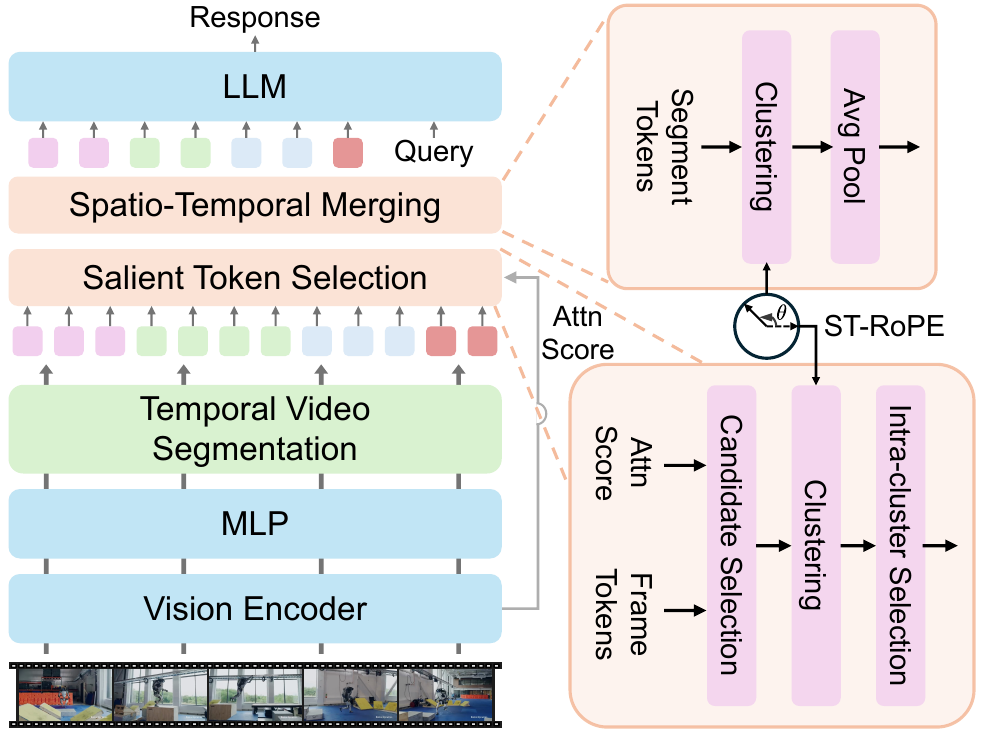}
\caption{\textbf{Overview of our method.}
It comprises three modules: Temporal Video Segmentation (TVS), Salient Token Selection (STS), and Spatio-Temporal Merging (STM).
Our technical contributions lie in (1) an enhanced salient token selection strategy in the \colorbox{main_block}{STS} module, and (2) position-aware clustering with ST-RoPE, used in \colorbox{main_block}{STS} and \colorbox{main_block}{STM} modules.
}

\label{fig:method_overview}
\end{figure}

\section{Our Method}
As shown in~\cref{fig:method_overview}, our pruning method comprises three modules: (1) temporal video segmentation, (2) salient token selection, and (3) spatio-temporal merging.
We detail these modules in the following sections.

\subsection{Temporal Video Segmentation}
Following~\cite{holitom}, we perform temporal video segmentation with the objective of maximizing prunable static tokens.
Let $h_{t,k}$ be the $k$-th vision token at frame $t$. 
A token is \textit{static} if its similarity between adjacent frames exceeds the threshold $\tau$ throughout the segment. 
For a segment $[t_s, t_e)$ starting at frame $t_s$ and ending at frame $t_e - 1$, the number of prunable tokens $g(t_s, t_e)$ is defined as:
\begin{equation}
\begin{split}
    g(t_s, t_e) &= \left( \sum_{k=1}^{N_v} \prod_{t=t_s}^{t_e-2} \mathbb{I}(\text{sim}(h_{t,k}, h_{t+1,k}) > \tau) \right) \\
    &\quad \times (t_e - t_s - 1)
    \text{.}
\end{split}
\end{equation}
To find the optimal segmentation boundaries $\{t_i\}$, a global maximum can be achieved via dynamic programming, with the state transition function:
\begin{equation}
    dp[i] = \max_{1 \le j < i} \{ dp[j] + g(j, i) \}
    \text{,}
\end{equation}
where $dp[i]$ denotes the maximum prunable tokens within the first $i-1$ frames, with the base case $dp[1]=0$.
Similar to previous approaches~\cite{holitom,fu2025framefusion}, we average-pool these prunable static tokens into the first frame of the segment.

\subsection{Salient Token Selection}

Given the observation that attention distribution is long-tailed and multimodal, the Top-$k$ strategy can be ineffective for capturing this pattern.
To address this issue, our high-level idea is to first expand the retention candidate set to better cover the tail of the distribution. 
Then, we perform clustering and intra-cluster selection. 
This helps cover diverse semantic regions when selecting salient tokens. 
We term this approach a \textit{diversity-driven selection method}.

Specifically, we first retrieve a broader set of top-$\bar{k}$ candidate token indices, denoted as $\mathcal{I}_c$, with $\bar{k}=\lfloor \alpha \cdot k \rfloor$, where $\alpha \geq 1$ is an expansion coefficient.  
Subsequently, to partition these candidates into $k$ distinct semantic regions, we employ Density Peaks Clustering based on $K$-Nearest Neighbors (DPC-KNN)~\cite{du2016study, rodriguez2014clustering}. 
For each candidate token $\bm{x}_i$, we compute its local density $\rho_i$, and $\delta_i$, its distance to the closest token with higher density:
\begin{align}
    \rho_i &= \exp\left(-\frac{1}{K} \sum_{\bm{x}_j \in \text{KNN}(\bm{x}_i)} \text{d}(\bm{x}_i, \bm{x}_j)^2 \right) \text{,} \notag \\
    \delta_i &= 
    \begin{cases} 
        \min\limits_{j \in \mathcal{I}_c: \rho_j > \rho_i} \text{d}(\bm{x}_i, \bm{x}_j), & \text{if } \exists j \in \mathcal{I}_c \text{ s.t. } \rho_j > \rho_i \\
        \max\limits_{j \in \mathcal{I}_c} \text{d}(\bm{x}_i, \bm{x}_j), & \text{otherwise}
    \end{cases}
    \text{,}
\end{align}
where $\text{KNN}(\bm{x}_i)$ denotes the set of $K$ nearest neighbors of $\bm{x}_i$, and $\text{d}(\cdot, \cdot)$ represents the Euclidean distance. 
The density score for each candidate is defined as $\gamma_i = \rho_i \times \delta_i$.  
Tokens with high $\gamma_i$ are selected as cluster centers. 
We then assign each remaining candidate to its nearest cluster center in the feature space, partitioning the candidate indices $\mathcal{I}_c$ into $k$ disjoint clusters $\{\mathcal{C}_c\}_{c=1}^k$.

Finally, we pick the token with the highest attention score per cluster. The final selected indices of salient tokens are
\begin{equation}
    \mathcal{I}_s^* = \bigcup_{c=1}^k \left\{ \underset{i \in \mathcal{C}_c}{\operatorname{arg\,max}} (\bar{A}_{i}) \right\} \text{.}
\end{equation}

\begin{figure}[!t]
\centering
 \setlength{\abovecaptionskip}{3mm}
    \setlength{\belowcaptionskip}{-2mm}
\includegraphics[width=0.98\columnwidth]{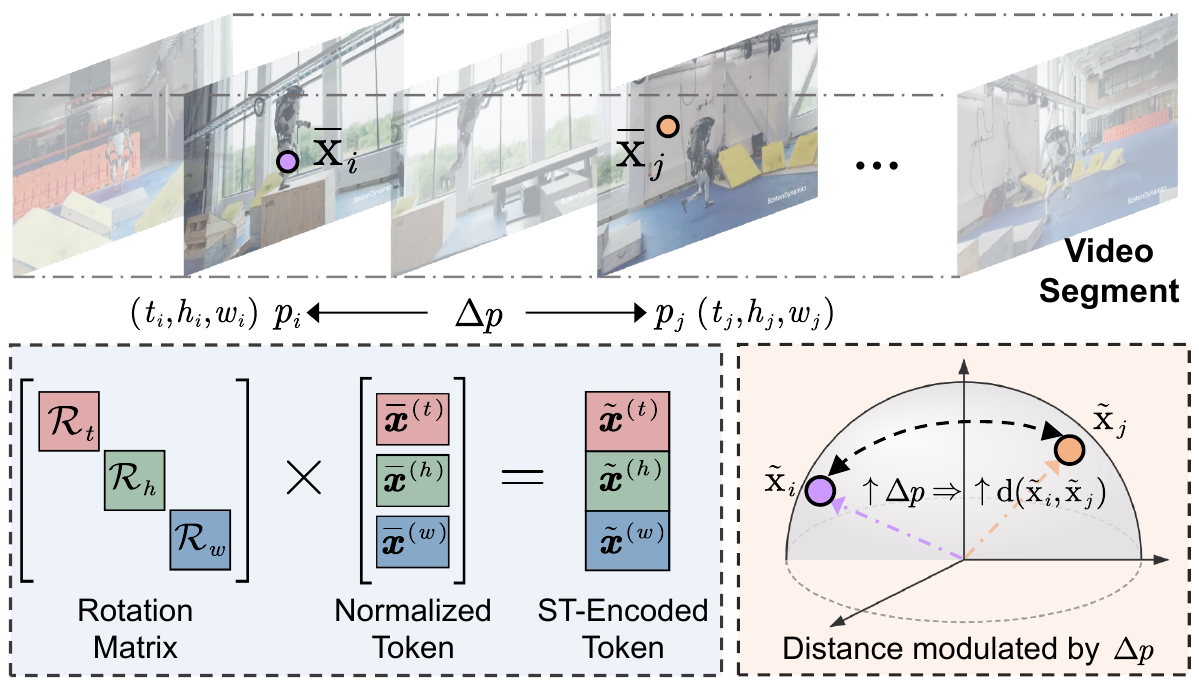}
\caption{\textbf{Illustration of ST-RoPE.}
(Bottom left) It encodes spatio-temporal position information by applying a rotation matrix to vision tokens.
(Bottom right) The pairwise distance on the hypersphere, $\text{d}(\cdot,\cdot)$, is not only determined by cosine similarity, but also modulated by $\Delta p$, where spatio-temporally distant tokens are assigned a larger distance penalty.
Consequently, semantically similar and spatio-temporally neighboring tokens are clustered together.}
\label{fig:st_rope}
\end{figure}

\subsection{Spatio-Temporal Merging}
Within each video segment, we aggregate similar tokens to reduce feature redundancy. 
More specifically, we use the DPC-KNN algorithm to group these tokens into separate clusters, as described in the previous section.
The vision tokens within each cluster are reduced with average pooling.

\paragraph{ST-RoPE.}
As previously introduced, the vanilla clustering approach leads to noisy representations.
To address this issue, we propose injecting spatio-temporal information coupled with a locality prior into the clustering process.
The key intuition is that, since objects are often continuous rather than fragmented, tokens of the same cluster should not only be semantically coherent but also spatio-temporally adjacent.
\cref{fig:st_rope} illustrates our high-level idea.

Specifically, we repurpose 1D RoPE~\cite{su2024roformer} as ST-RoPE by integrating spatio-temporal information.
Given two normalized tokens $\bar{\bm{x}}_i$ and $\bar{\bm{x}}_j$, their cosine similarity integrated with positional information is formulated as:
\begin{equation}
    \text{cos}_{\text{ST}}(\tilde{\bm{x}}_i, \tilde{\bm{x}}_j) = (\mathcal{R}_{\Theta, p_i}^d \bar{\bm{x}}_i)^\top(\mathcal{R}_{\Theta, p_j}^d \bar{\bm{x}}_j)= \bar{\bm{x}}_i^\top \mathcal{R}_{\Theta, \Delta p}^d \bar{\bm{x}}_j
    \text{,}
\end{equation}
where $\mathcal{R}_{\Theta, \Delta p}^d$ is a block-diagonal rotation matrix encoding the relative distance $\Delta p = p_j - p_i$. 
The 3D position is denoted by $p = (t, h, w)$, where $h \in \{1, \dots, H^\prime\}$ and $w \in \{1, \dots, W^\prime\}$ are spatial indices within the feature grid of a frame, and $t$ is the timestamp to robustly accommodate varying sampling strides.

To explicitly decouple the spatial and temporal dynamics, the pre-defined frequency parameter set $\Theta = \{ \Theta^{(t)}, \Theta^{(h)}, \Theta^{(w)} \}$ is formulated as:
\begin{equation}
    \theta_k^{(p)} = 
    \left\{
    \begin{array}{@{}l@{\quad}l@{\quad}r}
        \big(\theta_{\mathrm{base}}^{(t)}\big)^{\frac{-2(k-1)}{d_t}}, & p = t, & k \in[1, 2,\dots, d_t/2], \\[6pt]
        \big(\theta_{\mathrm{base}}^{(h)}\big)^{\frac{-2(k-1)}{d_h}}, & p = h, & k \in [1, 2,\dots, d_h/2], \\[6pt]
        \big(\theta_{\mathrm{base}}^{(w)}\big)^{\frac{-2(k-1)}{d_w}}, & p = w, & k \in [1, 2,\dots, d_w/2],
    \end{array}
    \right.
\end{equation}
where $\bar{\bm{x}} \in \mathbb{R}^d$ is partitioned as $\bar{x} =[\bar{\bm{x}}^{(t)}; \bar{\bm{x}}^{(h)}; \bar{\bm{x}}^{(w)}]$, with dimensions $d_t, d_h$, and $d_w$, respectively.

\paragraph{Locality Prior with Long-term Decay Property.}
A key property of our formulation is \textit{long-term decay}, which makes tokens that are far apart yield lower similarity scores.
To see this, consider the temporal section with a relative distance $m = t_j - t_i$. The inner product can be written as the real part of a complex multiplication:
\begin{equation}
    \bar{\bm{x}}_i^{(t)\top} \mathcal{R}_{\Theta^{(t)}, m} \bar{\bm{x}}_j^{(t)} = \text{Re} \left[ \sum_{k=0}^{d_t/2 - 1} c_k e^{\mathrm{i} m \theta_k^{(t)}} \right]
    \text{,}
\end{equation}
where $c_k = (\bar{\bm{x}}_{i, 2k}^{(t)} + \mathrm{i}\bar{\bm{x}}_{i, 2k+1}^{(t)})(\bar{\bm{x}}_{j, 2k}^{(t)} - \mathrm{i}\bar{\bm{x}}_{j, 2k+1}^{(t)})$, and $\mathrm{i}$ is the imaginary unit.
By denoting $S_l = \sum_{k=0}^{l-1} e^{\mathrm{i} m \theta_k^{(t)}}$, and letting $c_{d_t/2} = 0$ and $S_0 = 0$, we apply the Abel transformation to rewrite the sum, whose magnitude is upper-bounded by:
\begin{equation}
    \left| \sum_{k=0}^{d_t/2 - 1} c_k e^{\mathrm{i} m \theta_k^{(t)}} \right| \leq \left( \max_k |c_{k+1} - c_k| \right) \sum_{k=0}^{d_t/2 - 1} |S_{k+1}|
    \text{.}
\end{equation}
As shown in 1D RoPE formulation~\cite{su2024roformer,xu2024base}, the summation $\sum |S_{k+1}|$ decays as the relative distance $m$ increases.
This long-term decay property holds independently for the $t$, $h$, and $w$ sections. 
Thus, a greater relative distance in any spatio-temporal dimension will generally penalize the overall token-wise cosine similarity.

\paragraph{Compatibility with Clustering Metric.}
As noted previously, the DPC-KNN clustering algorithm often uses the Euclidean distance metric.
With normalized tokens, the Euclidean distance on the hypersphere can be reduced as
\begin{align}
    \| \bar{\bm{x}}_i - \bar{\bm{x}}_j \| 
    &= \sqrt{\|\bar{\bm{x}}_i\|^2 + \| \bar{\bm{x}}_j \|^2 - 2 \bar{\bm{x}}_i^\top \bar{\bm{x}}_j} \notag \\
    &= \sqrt{2 \cdot \big(1- \text{cos}(\bar{\bm{x}}_i, \bar{\bm{x}}_j) \big)}
    \text{,}
    \label{eq:l2_dist}
\end{align}
where a larger cosine similarity $\text{cos}(\bar{\bm{x}}_i, \bar{\bm{x}}_j)$ corresponds to smaller Euclidean distance.
Here, we show the compatibility of our formulation. 
Note that since the rotation matrices are orthogonal (\ie, $\mathcal{R}^\top \mathcal{R} = \mathbf{I}$), they preserve the $\ell_2$-norm of the vectors. 
Consequently, by substituting the normalized features with the rotated ones, \cref{eq:l2_dist} can be reformulated as the following equation:
\begin{equation}
    \| \tilde{\bm{x}}_i - \tilde{\bm{x}}_j \| =  \sqrt{2 \cdot \big( 1- \text{cos}_{\text{ST}}(\tilde{\bm{x}}_i, \tilde{\bm{x}}_j) \big)}
    \text{.}
\end{equation}
Since $\text{cos}_{\text{ST}}(\tilde{\bm{x}}_i, \tilde{\bm{x}}_j)$ is modulated by relative distance $\Delta p$, the distance metric enables grouping tokens that are both semantically similar and spatio-temporally neighboring.
\section{Experiments}

\subsection{Experimental Setting}
\paragraph{Benchmarks.} We conduct comprehensive experiments on four video understanding benchmarks. 
Specifically, the selected datasets comprise general open-world video understanding evaluations, including Video-MME~\cite{video-mme} and MVBench~\cite{mvbench}, as well as challenging benchmarks tailored for long-form video comprehension, namely LongVideoBench~\cite{longvideobench} and MLVU~\cite{mlvu}. 
Collectively, these datasets encompass a wide spectrum of domains, durations, and tasks, serving as an ideal testbed for validating the effectiveness and generalization capabilities of token compression methods.
More dataset details are provided in the Appendix.

\paragraph{Implementation Details.}
We implement and evaluate our method on three representative Video LLMs, LLaVA-OneVision-7B~\cite{llava-ov}, LLaVA-Video-7B~\cite{llava-video}, and Qwen2.5-VL-7B~\cite{qwen2.5-vl}.
More introductions are provided in the Appendix.
All experiments are performed on NVIDIA A800 GPUs with the VLMEvalKit~\cite{vlmevalkit} evaluation framework.
We use an extra global query to compute the attention score on tokens of each video frame~\cite{fastvid}, and mask constant attention sink tokens for salient token selection.

\paragraph{Hyperparameters.}
Unless otherwise specified, experiments are conducted with equivalent retention ratios of $\bar{r}=\{0.1,0.15,0.2\}$, and configure the $\text{KNN}$ algorithm to consider $7$ local neighbors.
For the temporal video segmentation phase, the similarity threshold is empirically set to $\tau=0.65$ when $\bar{r}=0.1$, and $0.8$ otherwise.
During salient token selection, we utilize an expansion coefficient of $\alpha = 1.5$. 
For spatio-temporal merging, the specific rotational configurations are defined as $d_t=1186, d_h=d_w=1184$, with base frequencies $\theta_{\mathrm{base}}^{(t)}=10^4, \theta_{\mathrm{base}}^{(h)}=\theta_{\mathrm{base}}^{(w)}=10^3$.
Regarding the total token retention budget, we allocate 60\% to salient token selection and 40\% to spatio-temporal merging.

\paragraph{Baselines.}
We compare our method against state-of-the-art training-free token pruning approaches. 
These include intra-LLM methods (FastV~\cite{fastv}, DART~\cite{dart}), pre-LLM methods (VisionZip~\cite{visionzip}, VidCom\textsuperscript{2}~\cite{vidcom2}, FastVID~\cite{fastvid}), and the hybrid method HoliTom~\cite{holitom}.
Further technical descriptions and implementation details are provided in the Appendix.
To ensure a fair comparison, all reported baseline results are reproduced under identical settings.

\begin{table*}[!t]
\centering
\footnotesize
\tablestyle{5.5pt}{1.2}
\setlength{\aboverulesep}{0.2ex}
\setlength{\belowrulesep}{0.2ex}

\caption{\textbf{Comparison of state-of-the-art token pruning methods on four mainstream video understanding benchmarks.} We experiment with three retention ratios (10\%, 15\%, 20\%) and 32 sampled frames on the LLaVA-OneVision-7B model. Our method consistently achieves the best overall performance across different retention ratios.
\textbf{Bold} and {\ul underlined} denote best and second-best performance, respectively. Avg.: Average accuracy on four benchmarks. Perf.: Ratio of performance retained relative to the original model.
All baseline results are reproduced under the same settings.
}
\label{tab:main_res}
\resizebox{0.96\linewidth}{!}{
\begin{tabular}{l !{\color{gray!60}\vrule width 0.3pt} ccccccc !{\color{gray!60}\vrule width 0.3pt} cr}
\toprule
\multirow{2}{*}{\textbf{Methods}} & \multicolumn{4}{c}{\textbf{Video-MME}~(w/o sub)}                              & \multirow{2}{*}{\textbf{MVBench}} & \multirow{2}{*}{\textbf{\makecell{LongVideo\\Bench}}} & \multirow{2}{*}{\textbf{MLVU}} & \multirow{2}{*}{\textbf{Avg.}} & \multirow{2}{*}{\textbf{Perf.}}    \\ \cmidrule(lr){2-5}
                                  & \textit{Short} & \textit{Medium} & \textit{Long} & \textit{Overall} &                                   &                                          &                                &                                &                                   \\ \midrule
\textcolor{grey}{LLaVA-OneVision-7B}~\cite{llava-ov}                       & \textcolor{grey}{71.9}           & \textcolor{grey}{56.3}            & \textcolor{grey}{49.6}          & \textcolor{grey}{59.3}             & \textcolor{grey}{55.3}                              & \textcolor{grey}{56.8}                                     & \textcolor{grey}{64.7}                           & \textcolor{grey}{59.0}                           & \textcolor{grey}{100.0}                             \\ \shline

\rowcolor{isabelline}
\multicolumn{10}{c}{\textit{Retention Ratio=10\%}}                                                                                                                                                                                                                                           \\ \shline
FastV~\cite{fastv}~\textsubscript{\textcolor{gray}{[ECCV 2024]}}                             & 54.7           & 49.0            & 43.7          & 49.1             & 48.3                              & 47.3                                     & 54.6                           & 49.8                           & 84.4                              \\
DART~\cite{dart}~\textsubscript{\textcolor{gray}{[EMNLP 2025]}}                              & 54.4           & 49.0            & 44.0          & 49.1             & 46.2                              & 46.3                                     & 55.4                           & 49.3                           & 83.5                              \\
VisionZip~\cite{visionzip}~\textsubscript{\textcolor{gray}{[CVPR 2025]}}                         & 61.2           & 52.3            & 47.2          & 53.6             & 50.7                              & 51.7                                     & 59.0                           & 53.8                           & 91.1                              \\
VidCom\textsuperscript{2}~\cite{vidcom2}~\textsubscript{\textcolor{gray}{[EMNLP 2025]}}                           & 61.7           & 51.7            & 47.7          & 53.7             & 49.4                              & 50.4                                     & 57.4                           & 52.7                           & 89.3                              \\
FastVID~\cite{fastvid}~\textsubscript{\textcolor{gray}{[NeurIPS 2025]}}                           & 65.1           & {\ul 55.3}      & {\ul 48.6}    & {\ul 56.3}       & 54.5                              & {\ul 55.7}                               & 61.2                           & 56.9                           & 96.4                              \\
HoliTom~\cite{holitom}~\textsubscript{\textcolor{gray}{[NeurIPS 2025]}}                           & {\ul 66.6}     & 53.8            & {\ul 48.6}    & {\ul 56.3}       & {\ul 54.9}                        & 55.4                                     & {\ul 61.6}                     & {\ul 57.1}                     & {\ul 96.7}                        \\

\rowcolor{lightblue}
\textbf{Tango} (Ours)                              & \textbf{68.9}  & \textbf{56.8}   & \textbf{49.8} & \textbf{58.5}    & \textbf{55.4}                     & \textbf{56.2}                            & \textbf{63.5}                  & \textbf{58.4}                  & \textbf{98.9}                     \\ \shline

\rowcolor{isabelline}
\multicolumn{10}{c}{\textit{Retention Ratio=15\%}}                                                                                                                                                                                                                                           \\ \shline
FastV~\cite{fastv}~\textsubscript{\textcolor{gray}{[ECCV 2024]}}
                             & 59.7           & 51.0            & 45.0          & 51.9             & 50.5                              & 50.8                                     & 57.3                           & 52.6                           & 89.1                              \\
VisionZip~\cite{visionzip}~\textsubscript{\textcolor{gray}{[CVPR 2025]}}
                         & 63.1           & 53.6            & 47.3          & 54.7             & 52.2                              & 52.4                                     & 59.8                           & 54.8                           & 92.8                              \\
VidCom\textsuperscript{2}~\cite{vidcom2}~\textsubscript{\textcolor{gray}{[EMNLP 2025]}}
                           & 66.3           & 54.8            & 48.2          & 56.4             & 52.6                              & 52.7                                     & 61.0                           & 55.7                           & 94.3                              \\
FastVID~\cite{fastvid}~\textsubscript{\textcolor{gray}{[NeurIPS 2025]}}
                           & {\ul 69.1}     & {\ul 55.4}      & {\ul 48.9}    & {\ul 57.8}       & {\ul 55.1}                        & 55.9                                     & {\ul 63.3}                     & {\ul 58.0}                     & {\ul 98.3}                        \\
HoliTom~\cite{holitom}~\textsubscript{\textcolor{gray}{[NeurIPS 2025]}}                           & 68.1           & 53.1            & 48.2          & 56.5             & \textbf{55.4}                     & \textbf{56.5}                            & 62.8                           & 57.8                           & 97.9                              \\
\rowcolor{lightblue}
\textbf{Tango} (Ours)                              & \textbf{69.8}  & \textbf{55.6}   & \textbf{49.7} & \textbf{58.3}    & \textbf{55.4}                     & {\ul 56.4}                               & \textbf{64.0}                  & \textbf{58.5}                  & \textbf{99.1}                     \\ \shline

\rowcolor{isabelline}
\multicolumn{10}{c}{\textit{Retention Ratio=20\%}}                                                                                                                                                                                                                                           \\ \shline
FastV~\cite{fastv}~\textsubscript{\textcolor{gray}{[ECCV 2024]}}
                             & 63.8           & 51.4            & 45.9          & 53.7             & 51.9                              & 54.7                                     & 59.0                           & 54.8                           & 92.9          \\
VisionZip~\cite{visionzip}~\textsubscript{\textcolor{gray}{[CVPR 2025]}}
                         & 65.1           & 55.1            & 47.9          & 56.0             & 53.8                              & 54.8                                     & 62.1                           & 56.7                           & 96.0          \\
VidCom\textsuperscript{2}~\cite{vidcom2}~\textsubscript{\textcolor{gray}{[EMNLP 2025]}}                           & 69.6           & {\ul 56.0}      & {\ul 49.3}    & 58.3             & 53.4                              & 53.7                                     & 62.0                           & 56.8                           & 96.3         \\
FastVID~\cite{fastvid}~\textsubscript{\textcolor{gray}{[NeurIPS 2025]}}
                           & 70.2           & 55.8            & \textbf{49.4} & {\ul 58.5}       & {\ul 55.5}                        & 56.2                                     & {\ul 63.7}                     & {\ul 58.5}                     & {\ul 99.0}    \\
HoliTom~\cite{holitom}~\textsubscript{\textcolor{gray}{[NeurIPS 2025]}}                           & {\ul 71.2}     & 53.8            & \textbf{49.4} & 58.1             & 55.3                              & {\ul 56.5}                               & 63.0                           & 58.2                           & 98.6          \\
\rowcolor{lightblue}
\textbf{Tango} (Ours)                              & \textbf{71.3}  & \textbf{56.1}   & 48.9          & \textbf{58.8}    & \textbf{55.9}                     & \textbf{56.7}                            & \textbf{64.1}                  & \textbf{58.9}                  & \textbf{99.7} \\
\bottomrule
\end{tabular}
}
\end{table*}

\subsection{Main Results}
In this section, we present the main results from our experiments, covering four aspects: 
(1) comparison with other SOTA methods; 
(2) cross-model generalizability; 
(3) compatibility with intra-LLM approaches;
and (4) scaling capability with more input frames.

\paragraph{Comparison with SOTAs.}
As shown in~\cref{tab:main_res}, our method outperforms previous SOTAs by a clear margin. The advantage is even clearer in low-token-budget settings.
For instance, with a 10\% retention ratio, our method still preserves 98.9\% of the original model's performance, surpassing HoliTom by 2.2\% and FastVID by 2.5\%.

\begin{table*}[!t]
\centering
\footnotesize
\tablestyle{5.5pt}{1.2}
\setlength{\aboverulesep}{0.2ex}
\setlength{\belowrulesep}{0.2ex}

\caption{\textbf{Cross-model evaluation on LLaVA-Video-7B and Qwen2.5-VL-7B}, with 64 and 32 input frames, respectively.
$\dagger$: Combined with intra-LLM method (FastV variant~\cite{holitom}), our method can achieve better overall performance.}
\label{tab:transfer-res}
\resizebox{0.96\linewidth}{!}{
\begin{tabular}{l !{\color{gray!60}\vrule width 0.3pt} ccccccc !{\color{gray!60}\vrule width 0.3pt} cr}
\toprule
\multirow{2}{*}{\textbf{Methods}} & \multicolumn{4}{c}{\textbf{Video-MME}~(w/o sub)}                              & \multirow{2}{*}{\textbf{MVBench}} & \multirow{2}{*}{\textbf{\makecell{LongVideo\\Bench}}} & \multirow{2}{*}{\textbf{MLVU}} & \multirow{2}{*}{\textbf{Avg.}} & \multirow{2}{*}{\textbf{Perf.}}    \\ \cmidrule(lr){2-5}
                                  & \textit{Short} & \textit{Medium} & \textit{Long} & \textit{Overall} &                                   &                                          &                                &                                &                                   \\ \midrule
\textcolor{grey}{LLaVA-Video-7B}~\cite{llava-video}      & \textcolor{grey}{76.7} & \textcolor{grey}{62.0} & \textcolor{grey}{51.4} & \textcolor{grey}{63.4} & \textcolor{grey}{60.7} & \textcolor{grey}{57.5} & \textcolor{grey}{71.4} & \textcolor{grey}{63.3}                     & \textcolor{grey}{100.0} \\ 
\shline
\rowcolor{isabelline}
\multicolumn{10}{c}{\textit{Retention Ratio=10\%}}                                                                                                                        \\ 
\shline
FastV~\cite{fastv}~\textsubscript{\textcolor{gray}{[ECCV 2024]}}
               & 56.1           & 49.6            & 44.4          & 50.0             & 47.6          & 45.8          & 52.2          & 48.9          & 77.3          \\
VisionZip~\cite{visionzip}~\textsubscript{\textcolor{gray}{[CVPR 2025]}}           & 67.0           & 54.6            & 46.8          & 56.1             & 54.2          & 52.4          & 62.9          & 56.4          & 89.1          \\
VidCom\textsuperscript{2}~\cite{vidcom2}~\textsubscript{\textcolor{gray}{[EMNLP 2025]}}
             & 65.6           & 55.7            & 45.9          & 55.7             & 52.1          & 50.4          & 59.7          & 54.5          & 86.1          \\
FastVID~\cite{fastvid}~\textsubscript{\textcolor{gray}{[NeurIPS 2025]}}
             & 71.7           & 58.2            & 48.7          & 59.5             & \textbf{58.2} & 53.2          & 65.1          & 59.0          & 93.3          \\
HoliTom~\cite{holitom}~\textsubscript{\textcolor{gray}{[NeurIPS 2025]}}             & 73.1           & 58.8            & 48.7          & {\ul 60.2}       & 57.5          & {\ul 54.2}    & 66.2          & 59.5          & 94.1          \\
\rowcolor{lightblue}
\textbf{Tango} (Ours)               & 73.4           & {\ul 59.3}      & \textbf{49.4} & \textbf{60.7}    & 56.9          & {\ul 54.2}    & {\ul 66.7}    & {\ul 59.6}    & {\ul 94.2}    \\
\rowcolor{lightblue}
\textbf{Tango}$^{\dagger}$ (w/ intra-LLM) & 73.1           & \textbf{59.6}   & {\ul 49.3}    & \textbf{60.7}    & {\ul 57.6}    & \textbf{54.7} & \textbf{67.6} & \textbf{60.2} & \textbf{95.1} \\ 
\shline
\rowcolor{isabelline}
\multicolumn{10}{c}{\textit{Retention Ratio=20\%}}                                                                                                                        \\ 
\shline
VisionZip~\cite{visionzip}~\textsubscript{\textcolor{gray}{[CVPR 2025]}}           & 73.1           & 59.0            & 48.9          & 60.3             & 58.2          & 54.6          & 67.7          & 60.2          & 95.2          \\
FastVID~\cite{fastvid}~\textsubscript{\textcolor{gray}{[NeurIPS 2025]}}
             & 74.8           & 60.6            & 50.2          & 61.9             & \textbf{59.1} & 54.5          & 68.1          & 60.9          & 96.3          \\
HoliTom~\cite{holitom}~\textsubscript{\textcolor{gray}{[NeurIPS 2025]}}             & 74.7           & 60.9            & {\ul 50.3}    & {\ul 62.0}       & 58.7          & 54.2          & 68.2          & 60.8          & 96.1          \\
\rowcolor{lightblue}
\textbf{Tango} (Ours)                & {\ul 74.9}     & \textbf{61.7}   & 49.3          & {\ul 62.0}       & 58.5          & \textbf{55.6} & {\ul 68.7}    & {\ul 61.2}    & {\ul 96.8}    \\
\rowcolor{lightblue}
\textbf{Tango}$^{\dagger}$ (w/ intra-LLM) &
  \textbf{75.3} &
  {\ul 61.1} &
  \textbf{50.4} &
  \textbf{62.3} &
  {\ul 58.9} &
  {\ul 55.4} &
  \textbf{70.4} &
  \textbf{61.7} &
  \textbf{97.6} \\

\midrule
\midrule
\textcolor{grey}{Qwen2.5-VL-7B}~\cite{qwen2.5-vl}       & \textcolor{grey}{74.0}           & \textcolor{grey}{62.7}            & \textcolor{grey}{52.4}          & \textcolor{grey}{63.0}             & \textcolor{grey}{65.7}          & \textcolor{grey}{59.9}          & \textcolor{grey}{64.4}          & \textcolor{grey}{63.2}          & \textcolor{grey}{100.0}         \\
\shline
\rowcolor{isabelline}
\multicolumn{10}{c}{\textit{Retention Ratio=20\%}}                                                                                                                       \\
\shline
VisionZip~\cite{visionzip}~\textsubscript{\textcolor{gray}{[CVPR 2025]}}           & {\ul 69.8}     & 56.7            & 49.1          & 58.5             & {\ul 61.9}    & {\ul 55.7}    & {\ul 60.6}    & {\ul 59.2}    & {\ul 93.6}    \\
FastVID~\cite{fastvid}~\textsubscript{\textcolor{gray}{[NeurIPS 2025]}}
             & 68.9           & {\ul 56.8}      & {\ul 50.3}    & {\ul 58.7}       & 61.2          & \textbf{55.9} & 60.3          & 59.0          & 93.3          \\
\rowcolor{lightblue}
\textbf{Tango} (Ours)               & \textbf{70.2}  & \textbf{59.1}   & \textbf{50.6} & \textbf{60.0}    & \textbf{62.4} & \textbf{55.9} & \textbf{61.0} & \textbf{59.8} & \textbf{94.6} \\
\bottomrule
\end{tabular}
}
\end{table*}
\paragraph{Cross-Model Generalizability.}
Results on LLaVA-Video-7B and Qwen2.5-VL-7B are shown in~\cref{tab:transfer-res}.
Our method consistently outperforms previous SOTAs by a clear margin.
With a 20\% retention ratio, our approach surpasses VidCom\textsuperscript{2} by 4.6\% on LLaVA-Video-7B, and outperforms FastVID by 1.3\% on Qwen2.5-VL-7B.

\paragraph{Compatibility with Intra-LLM Method.}
On LLaVA-Video-7B, we experiment with combining an intra-LLM pruning method~\cite{holitom}, and perform an additional 50\% pruning after the $18$-th layer of the LLM.
As shown in~\cref{tab:transfer-res}, the approach achieves gains of 0.9 and 0.8 points under 10\% and 20\% retention ratios, respectively. 
The gains can be attributed to two factors: 
(1) Pruning at deep layers of LLM barely hurts performance, even under radical pruning ratios such as 50\%~\cite{holitom,fastv};
(2) Combining intra-LLM pruning enables a larger pre-LLM retention ratio and thus better performance.
Specifically, given that $\bar{r}=\bar{r}_{pre}\times \bar{r}_{intra}$ ($\bar{r}_{pre}$ and $\bar{r}_{intra}$ are retention ratios for pre- and intra-LLM methods), we can set $\bar{r}_{pre}=12.2\%>\bar{r}=10\%$.

\begin{figure}[!th]\centering
	\includegraphics[width=0.9\columnwidth]{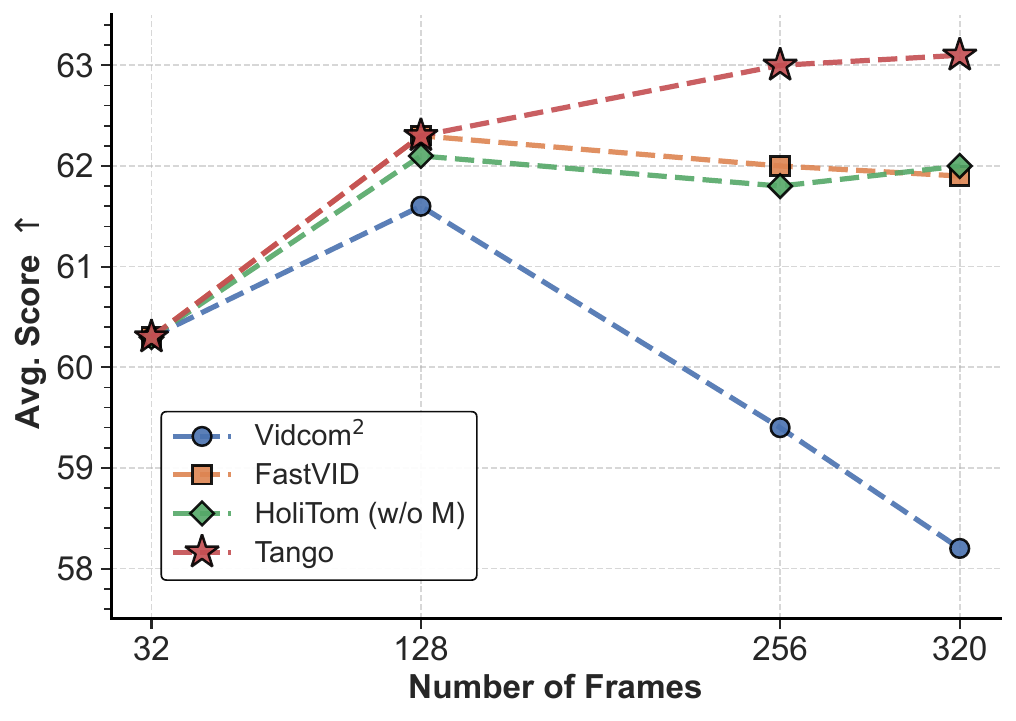}
	\caption{\textbf{Results with different input frames.}
    Our method maintains stable scaling with frame number, clearly outperforming other SOTA methods.
    ``w/o M'' denotes pre-LLM-only pruning~\cite{holitom}.
    The performance is averaged across three video datasets: Video-MME, LongVideoBench, and MLVU.}
	\label{fig:scale_frames}
\end{figure}

\paragraph{Scaling with Input Frames.} \cref{fig:scale_frames} compares different methods in terms of frame scalability (32$\times$196 token limit). 
As frame sampling becomes denser, spatio-temporal redundancy becomes more pronounced. 
Our method scales robustly with the frame count and consistently outperforms SOTA methods, demonstrating superior capability to retain essential spatio-temporal information.

\subsection{Ablation Study}
We perform an ablation study on LLaVA-OneVision-7B with a default retention ratio $\bar{r}=10\%$.

\begin{figure}[!th]\centering
\setlength{\abovecaptionskip}{2mm}
\setlength{\belowcaptionskip}{-4mm}
	\includegraphics[width=0.92\columnwidth]{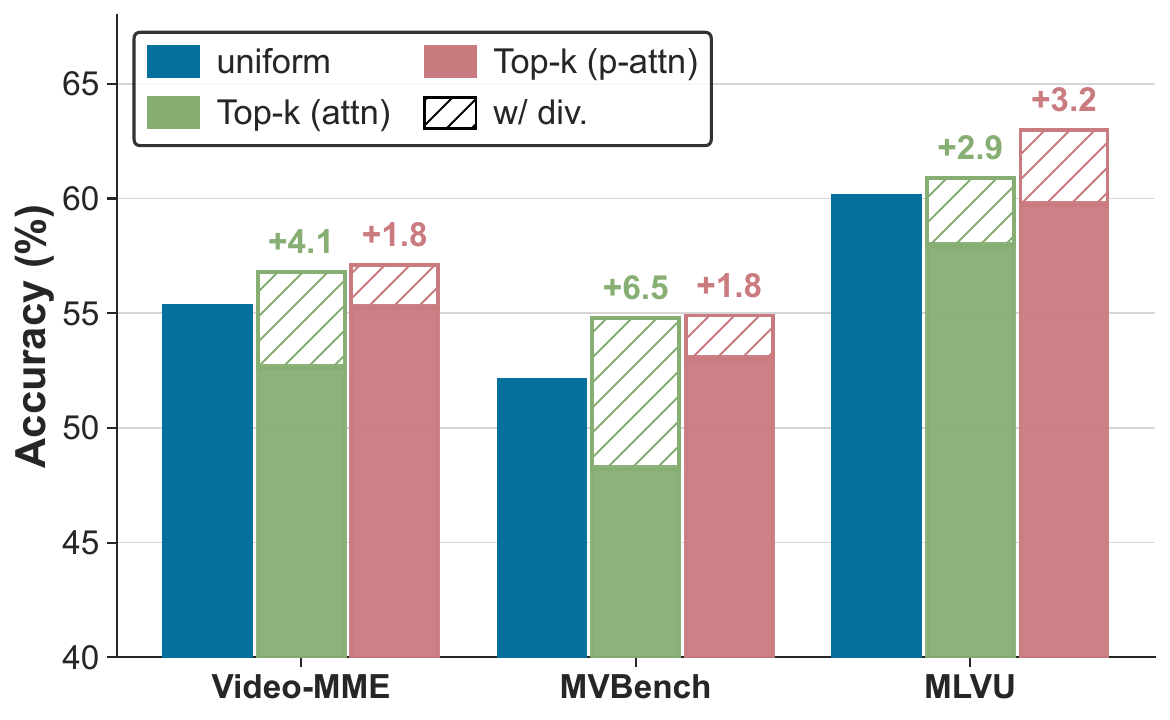}
 \caption{\textbf{Results with different salient token selection strategy}.
    Top-k (attn) underperforms, and Top-k (p-attn) is only comparable to a simple uniform sampling baseline. 
    Plugging in our proposed diversity-driven selection method (``w/ div.'') brings remarkable gains.
    \colorbox{attn_color}{\textcolor{white}{attn}} and \colorbox{p-attn_color}{\textcolor{white}{p-attn}} denote attention scores extracted by pooling attention weights on the query dimension~\cite{visionzip,holitom} and calculating with a global query~\cite{fastvid}, respectively.}
	\label{fig:abla_prune_metric}
\end{figure}

\paragraph{Salient Token Selection Strategy.}
\cref{fig:abla_prune_metric} compares three variants of salient token selection strategies: uniform sampling, Top-$k$ selection with ``attn''~\cite{visionzip,holitom}, and ``p-attn''~\cite{fastvid} scores.
We observe that vanilla Top-$k$ selection can be suboptimal: the ``attn'' variant even lags behind a simple uniform sampling baseline (52.7\% \vs 55.3\% on Video-MME).
Moreover, while ``p-attn'' generally performs better than ``attn'' (see Appendix for analyses on why ``p-attn'' is superior), it only matches the performance of uniform sampling.

Notably, incorporating our proposed diversity-driven strategy (``w/ div.'') yields significant gains of 6.5\% and 1.8\% on MVBench with ``attn'' and ``p-attn'' scores, respectively.

\begin{table}[!th]
\centering
\footnotesize
\tablestyle{5pt}{1.15}
\setlength{\aboverulesep}{0.2ex}
\setlength{\belowrulesep}{0.2ex}

\caption{\textbf{Ablation study of the proposed components.} w/ div.: Diversity-driven salient token selection strategy. ST Merge: Spatio-temporal merging. TSA: Timestamp alignment for each video frame, otherwise a fixed increment is used.}
\label{tab:ablation}
\begin{tabular}{cccc !{\color{gray!60}\vrule width 0.3pt} ccc}
\toprule
w/ div. & ST Merge & ST-RoPE & TSA & V-MME & LVB & Perf. \\ \midrule
\rowcolor{Gray!20}
\multicolumn{7}{l}{\textit{Effect on Token Selection}}    \\
\xmark       & \xmark        & \xmark       & \xmark   & 55.3  & 50.7 & 92.7 \\
\cmark       & \xmark        & \xmark       & \xmark   & 57.1  & 54.1 & 97.0 \\
\midrule
\rowcolor{Gray!20}
\multicolumn{7}{l}{\textit{Effect on ST Merging}}       \\
\xmark       & \cmark        & \xmark       & \xmark   & 56.9  & 54.2 & 96.7 \\
\xmark       & \cmark        & \cmark       & \xmark   & 57.3  & 55.1 & 97.3 \\
\xmark       & \cmark        & \cmark       & \cmark   & 57.9  & 55.8 & 98.1 \\
\midrule
\rowcolor{Gray!20}
\multicolumn{7}{l}{\textit{Full Method}}                 \\
\cmark       & \cmark        & \cmark       & \cmark   & 58.5  & 56.2 & 98.9 \\ \bottomrule
\end{tabular}
\end{table}
\paragraph{Ablation of Designed Components.}
Ablation results of our designed components are summarized in~\cref{tab:ablation}.

\textit{Improving attention-based salient token selection}: 
Compared with direct Top-$k$ selection, our strategy yields a significant overall gain of 5.3\%.
The results indicate the superiority of our proposed strategy and the importance of covering diverse semantic clusters when selecting salient tokens.

\textit{Improving similarity-based spatio-temporal merging}:
Incorporating the ST-RoPE mechanism improves overall accuracy by 0.6\%, indicating the effectiveness of injecting spatio-temporal locality prior.
Moreover, adding timestamp-aligned design brings another 0.8\% overall improvement.
Compared with MRoPE~\cite{qwen2-vl}, which uses a fixed temporal increment, this design more adaptively accommodates varying video lengths and sampling strides.
Finally, the assembly of all components yields the best overall performance, indicating the synergy of the designed components.

\begin{figure}[!t]\centering
\setlength{\abovecaptionskip}{2mm}
	\includegraphics[width=0.95\columnwidth]{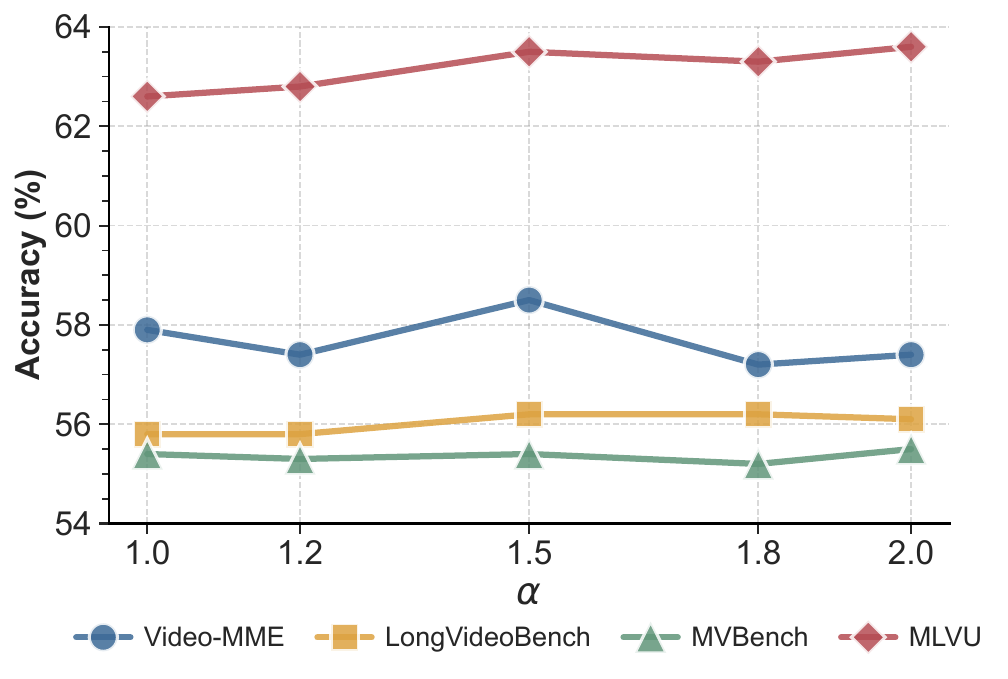}
    \caption{\textbf{Impact of expansion coefficient.} Overall accuracy is relatively stable \wrt varying coefficients.
    Since excess candidate tokens can also introduce background noise, a moderate value ($1.5$) yields the overall best results.}
	\label{fig:abla_alpha}
\end{figure}

\begin{table}[!t]
\centering
\footnotesize
\tablestyle{5.5pt}{1.2}
\setlength{\aboverulesep}{0.2ex}
\setlength{\belowrulesep}{0.2ex}

\caption{\textbf{Impact of rotational base frequencies.} Smaller frequencies induce a relatively larger decay effect. The best configuration is \colorbox{lightblue}{highlighted}. Note that $\theta_{\mathrm{base}}^{(s)}=\theta_{\mathrm{base}}^{(h)}=\theta_{\mathrm{base}}^{(w)}$.}
\label{tab:abla_theta}
\begin{tabular}{cc !{\color{gray!60}\vrule width 0.3pt} ccccc}
\toprule
\multirow{2}{*}{$\theta_{\mathrm{base}}^{(t)}$} & \multirow{2}{*}{$\theta_{\mathrm{base}}^{(s)}$} & \multicolumn{4}{c}{Video-MME}                                       & \multirow{2}{*}{MLVU} \\ \cmidrule(lr){3-6}
                          &                           & \textit{Short} & \textit{Medium} & \textit{Long} & \textit{Overall} &                       \\ 
\midrule
1e4 & 1e2 & 67.8 & 56.0 & 49.7 & 57.8 & 63.3 \\
\rowcolor{lightblue}
1e4 & 1e3 & 68.9 & 56.8 & 49.8 & 58.5 & 63.5 \\
1e4 & 1e4 & 67.6 & 56.4 & 48.9 & 57.6 & 63.3 \\
1e3 & 1e3 & 67.2 & 56.0 & 49.1 & 57.4 & 63.4 \\
1e5 & 1e3 & 67.4 & 55.7 & 49.9 & 57.7 & 63.4 \\ \bottomrule
\end{tabular}
\vspace{-2mm}
\end{table}
\paragraph{Hyperparameter Analysis.}
We analyze the impact of two critical hyperparameters, the expansion coefficient in the STM module and the base frequencies for ST-RoPE.

\cref{fig:abla_alpha} shows the ablation results of varying the expansion coefficient $\alpha$.
Expanding the candidate retained set allows more semantic regions to be included. However, incorporating too many tokens can introduce background noise. 
Empirical results indicate that a moderate value ($1.5$) optimally balances this trade-off.

\cref{tab:abla_theta} shows results of different configurations of base frequencies $\theta_{\mathrm{base}}$ in temporal and spatial dimensions.
Generally, a smaller frequency induces a larger decay effect, thereby a stronger locality prior. Yet, an excessively strong prior may disrupt the semantic similarity relationships.
Consequently, we adopt a moderate configuration that provides the best overall results.

\begin{figure}[!t]\centering
\setlength{\abovecaptionskip}{2mm}
\setlength{\belowcaptionskip}{-4mm}
	\includegraphics[width=\columnwidth]{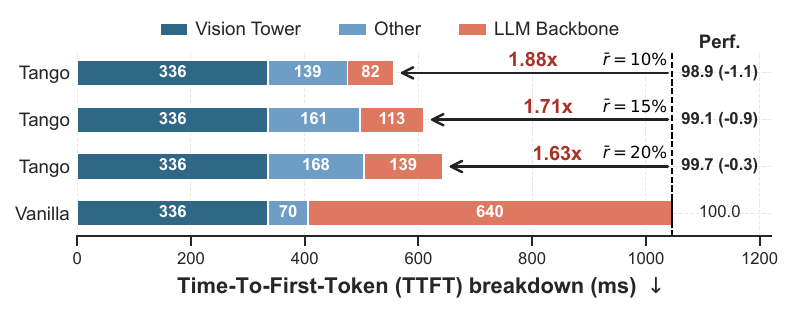}
 \caption{\textbf{Latency and relative performance under different retention ratios.}
Our method maintains 99.7\% performance relative to the full-token setting, with a 1.63$\times$ speedup.
``Other'' includes the projector, pooling, and additional pruning-related computations.
Wall-clock time is measured on the Video-MME benchmark.}
	\label{fig:effi}
\end{figure}

\subsection{Efficiency Analysis}
\cref{fig:effi} illustrates the latency-performance trade-off of our method. Our approach strikes an exceptional balance between efficiency and accuracy: at 10\% token retention, it preserves 98.9\% of full-token performance with a $1.88\times$ speedup. Increasing retention to 20\% yields nearly lossless performance (99.7\%) and a $1.63\times$ speedup.

\section{Conclusion}
\label{sec:conclusion}
In this paper, we revisit and advance two foundational token reduction paradigms: attention-based token selection and similarity-based token merging. 
Our proposed method, Tango, introduces diversity-driven token selection and explicitly incorporates a spatio-temporal locality prior into the merging process. 
Extensive experiments demonstrate the general effectiveness, input scalability, and highly favorable efficiency-performance trade-offs of our approach. 
Notably, Tango preserves 98.9\% of the full-token performance while accelerating inference by 1.88$\times$, establishing it as a highly practical and robust optimization technique for Video LLMs.

{\small
\bibliographystyle{unsrtnat}
\bibliography{references}

@String(CVPR= {IEEE Conf. Comput. Vis. Pattern Recog.})

@String(ICCV= {Int. Conf. Comput. Vis.})

@String(ECCV= {Eur. Conf. Comput. Vis.})

@String(ICLR = {Int. Conf. Learn. Represent.})

@String(CVPR  = {CVPR})

@String(ICCV  = {ICCV})

@String(ECCV  = {ECCV})

@String(ICLR  = {ICLR})

@article{qwen2.5-vl,
  title={{Qwen2.5-VL Technical Report}},
  author={Bai, Shuai and Chen, Keqin and Liu, Xuejing and Wang, Jialin and Ge, Wenbin and Song, Sibo and Dang, Kai and Wang, Peng and Wang, Shijie and Tang, Jun and others},
  journal={arXiv:2502.13923},
  year={2025}
}

@inproceedings{fastv,
  title={{An Image is Worth 1/2 Tokens After Layer 2: Plug-and-Play Inference Acceleration for Large Vision-Language Models}},
  author={Chen, Liang and Zhao, Haozhe and Liu, Tianyu and Bai, Shuai and Lin, Junyang and Zhou, Chang and Chang, Baobao},
  booktitle={ECCV},
  year={2024}
}

@inproceedings{dart,
  title={{Stop Looking for “Important Tokens” in Multimodal Language Models: Duplication Matters More}},
  author={Wen, Zichen and Gao, Yifeng and Wang, Shaobo and Zhang, Junyuan and Zhang, Qintong and Li, Weijia and He, Conghui and Zhang, Linfeng},
  booktitle={EMNLP},
  year={2025}
}

@inproceedings{divprune,
  title={{DivPrune: Diversity-based Visual Token Pruning for Large Multimodal Models}},
  author={Alvar, Saeed Ranjbar and Singh, Gursimran and Akbari, Mohammad and Zhang, Yong},
  booktitle={CVPR},
  year={2025}
}

@inproceedings{fu2025framefusion,
  title={{FrameFusion: Combining Similarity and Importance for Video Token Reduction on Large Vision Language Models}},
  author={Fu, Tianyu and Liu, Tengxuan and Han, Qinghao and Dai, Guohao and Yan, Shengen and Yang, Huazhong and Ning, Xuefei and Wang, Yu},
  booktitle={ICCV},
  year={2025}
}

@inproceedings{visionzip,
  title={{VisionZip: Longer is Better but Not Necessary in Vision Language Models}},
  author={Yang, Senqiao and Chen, Yukang and Tian, Zhuotao and Wang, Chengyao and Li, Jingyao and Yu, Bei and Jia, Jiaya},
  booktitle={CVPR},
  year={2025}
}

@inproceedings{vidcom2,
  title={{Video Compression Commander: Plug-and-Play Inference Acceleration for Video Large Language Models}},
  author={Liu, Xuyang and Wang, Yiyu and Ma, Junpeng and Zhang, Linfeng},
  booktitle={EMNLP},
  year={2025}
}

@article{yin2024survey,
  title={A survey on multimodal large language models},
  author={Yin, Shukang and Fu, Chaoyou and Zhao, Sirui and Li, Ke and Sun, Xing and Xu, Tong and Chen, Enhong},
  journal={National Science Review},
  year={2024}
}

@inproceedings{video-mme,
  title={{Video-MME: The First-Ever Comprehensive Evaluation Benchmark of Multi-modal LLMs in Video Analysis}},
  author={Fu, Chaoyou and Dai, Yuhan and Luo, Yongdong and Li, Lei and Ren, Shuhuai and Zhang, Renrui and Wang, Zihan and Zhou, Chenyu and Shen, Yunhang and Zhang, Mengdan and others},
  booktitle={CVPR},
  year={2025}
}

@inproceedings{mvbench,
  title={{MVBench: A Comprehensive Multi-modal Video Understanding Benchmark}},
  author={Li, Kunchang and Wang, Yali and He, Yinan and Li, Yizhuo and Wang, Yi and Liu, Yi and Wang, Zun and Xu, Jilan and Chen, Guo and Luo, Ping and others},
  booktitle={CVPR},
  year={2024}
}

@inproceedings{longvideobench,
  title={{LongVideoBench: A Benchmark for Long-context Interleaved Video-Language Understanding}},
  author={Wu, Haoning and Li, Dongxu and Chen, Bei and Li, Junnan},
  booktitle={NeurIPS},
  year={2024}
}

@inproceedings{mlvu,
  title={{MLVU: Benchmarking Multi-task Long Video Understanding}},
  author={Zhou, Junjie and Shu, Yan and Zhao, Bo and Wu, Boya and Liang, Zhengyang and Xiao, Shitao and Qin, Minghao and Yang, Xi and Xiong, Yongping and Zhang, Bo and others},
  booktitle={CVPR},
  year={2025}
}

@article{llava-ov,
  title={{LLaVA-OneVision: Easy Visual Task Transfer}},
  author={Li, Bo and Zhang, Yuanhan and Guo, Dong and Zhang, Renrui and Li, Feng and Zhang, Hao and Zhang, Kaichen and Zhang, Peiyuan and Li, Yanwei and Liu, Ziwei and others},
  journal={Transactions on Machine Learning Research},
  year={2025}
}

@article{llava-video,
  title={{LLaVA-Video: Video Instruction Tuning With Synthetic Data}},
  author={Zhang, Yuanhan and Wu, Jinming and Li, Wei and Li, Bo and Ma, Zejun and Liu, Ziwei and Li, Chunyuan},
  journal={Transactions on Machine Learning Research},
  year={2025}
}

@inproceedings{pyramiddrop,
  title={{PyramidDrop: Accelerating Your Large Vision-Language Models via Pyramid Visual Redundancy Reduction}},
  author={Xing, Long and Huang, Qidong and Dong, Xiaoyi and Lu, Jiajie and Zhang, Pan and Zang, Yuhang and Cao, Yuhang and He, Conghui and Wang, Jiaqi and Wu, Feng and others},
  booktitle={CVPR},
  year={2025}
}

@inproceedings{sparsevlm,
  title={{SparseVLM: Visual Token Sparsification for Efficient Vision-Language Model Inference}},
  author={Zhang, Yuan and Fan, Chun-Kai and Ma, Junpeng and Zheng, Wenzhao and Huang, Tao and Cheng, Kuan and Gudovskiy, Denis and Okuno, Tomoyuki and Nakata, Yohei and Keutzer, Kurt and others},
  booktitle={ICML},
  year={2025}
}

@article{qwen2,
  title={{Qwen2 Technical Report}},
  author={Yang, An and Yang, Baosong and Hui, Binyuan and Zheng, Bo and Yu, Bowen and Zhou, Chang and Li, Chengpeng and Li, Chengyuan and Liu, Dayiheng and Huang, Fei and others},
  journal={arXiv:2407.10671},
  year={2024}
}

@inproceedings{prunevid,
  title={{PruneVid: Visual Token Pruning for Efficient Video Large Language Models}},
  author={Huang, Xiaohu and Zhou, Hao and Han, Kai},
  booktitle={ACL (Findings)},
  year={2025}
}

@inproceedings{holitom,
  title={{HoliTom: Holistic Token Merging for Fast Video Large Language Models}},
  author={Shao, Kele and Tao, Keda and Qin, Can and You, Haoxuan and Sui, Yang and Wang, Huan},
  booktitle={NeurIPS},
  year={2025}
}

@inproceedings{fastvid,
  title={{FastVID: Dynamic Density Pruning for Fast Video Large Language Models}},
  author={Shen, Leqi and Gong, Guoqiang and He, Tao and Zhang, Yifeng and Liu, Pengzhang and Zhao, Sicheng and Ding, Guiguang},
  booktitle={NeurIPS},
  year={2025}
}

@article{du2016study,
  title={{Study on density peaks clustering based on k-nearest neighbors and principal component analysis}},
  author={Du, Mingjing and Ding, Shifei and Jia, Hongjie},
  journal={Knowledge-Based Systems},
  year={2016}
}

@article{rodriguez2014clustering,
  title={{Clustering by fast search and find of density peaks}},
  author={Rodriguez, Alex and Laio, Alessandro},
  journal={Science},
  year={2014}
}

@article{su2024roformer,
  title={{RoFormer: Enhanced transformer with Rotary Position Embedding}},
  author={Su, Jianlin and Ahmed, Murtadha and Lu, Yu and Pan, Shengfeng and Bo, Wen and Liu, Yunfeng},
  journal={Neurocomputing},
  year={2024}
}

@inproceedings{xu2024base,
  title={{Base of RoPE Bounds Context Length}},
  author={Xu, Mingyu and Men, Xin and Wang, Bingning and Zhang, Qingyu and Lin, Hongyu and Han, Xianpei and others},
  booktitle={NeurIPS},
  year={2024}
}

@article{qwen2-vl,
  title={{Qwen2-VL: Enhancing Vision-Language Model's Perception of the World at Any Resolution}},
  author={Wang, Peng and Bai, Shuai and Tan, Sinan and Wang, Shijie and Fan, Zhihao and Bai, Jinze and Chen, Keqin and Liu, Xuejing and Wang, Jialin and Ge, Wenbin and others},
  journal={arXiv:2409.12191},
  year={2024}
}

@inproceedings{siglip,
  title={{Sigmoid Loss for Language Image Pre-Training}},
  author={Zhai, Xiaohua and Mustafa, Basil and Kolesnikov, Alexander and Beyer, Lucas},
  booktitle={ICCV},
  year={2023}
}

@inproceedings{jiang2025vision,
  title={{Vision Transformers Don't Need Trained Registers}},
  author={Jiang, Nick and Dravid, Amil and Efros, Alexei and Gandelsman, Yossi},
  booktitle={NeurIPS},
  year={2025}
}

@inproceedings{darcet2023vision,
  title={{Vision Transformers Need Registers}},
  author={Darcet, Timoth{\'e}e and Oquab, Maxime and Mairal, Julien and Bojanowski, Piotr},
  booktitle={ICLR},
  year={2024}
}

@inproceedings{vlmevalkit,
  title={{VLMEvalKit: An Open-Source ToolKit for Evaluating Large Multi-Modality Models}},
  author={Duan, Haodong and Yang, Junming and Qiao, Yuxuan and Fang, Xinyu and Chen, Lin and Liu, Yuan and Dong, Xiaoyi and Zang, Yuhang and Zhang, Pan and Wang, Jiaqi and others},
  booktitle={ACM MM},
  year={2024}
}

@inproceedings{caba2015activitynet,
  title={{ActivityNet: A Large-Scale Video Benchmark for Human Activity Understanding}},
  author={Caba Heilbron, Fabian and Escorcia, Victor and Ghanem, Bernard and Carlos Niebles, Juan},
  booktitle={CVPR},
  year={2015}
}

@inproceedings{sun2024massive,
  title={Massive activations in large language models},
  author={Sun, Mingjie and Chen, Xinlei and Kolter, J Zico and Liu, Zhuang},
  booktitle={COLM},
  year={2024}
}
}

\clearpage 
\appendix 
\maketitlesupplementary
    
\section{Experimental Details}

\subsection{Benchmark Details}
\label{sec:bench_detail}

\paragraph{Video-MME}~\cite{video-mme} is a video understanding benchmark designed specifically for Video LLMs. It comprises a total of 900 videos and 2,700 multiple-choice questions. The video durations range from a few minutes to 1 hour.

\paragraph{MVBench}~\cite{mvbench} defines 20 video understanding tasks that encompass both spatial and temporal understanding.
The benchmark contains 4,000 multiple-choice questions.

\paragraph{LongVideoBench}~\cite{longvideobench} includes 3,763 web-scraped videos up to 1 hour long.
It curates 6,678 questions across 17 types of referring reasoning questions, challenging models' perception and relation reasoning capabilities.
We report results on the evaluation subset, which contains 1,337 questions.

\paragraph{MLVU}~\cite{mlvu} contains 1,730 videos spanning from 3 minutes to 2 hours, along with 3,102 QA pairs.
It comprises 9 task categories, including action understanding and topic reasoning.
We report results on the MLVU$_\texttt{M}$-Dev set.

\subsection{Model Details}
\label{sec:model_detail}

\paragraph{LLaVA-OneVision-7B}~\cite{llava-ov} is a representative fully open-sourced MLLM that unifies the understanding of single-image, multi-image, and video inputs.
We sample 32 frames for each video, resulting in 32$\times$196 video tokens.

\paragraph{LLaVA-Video-7B}~\cite{llava-video} is a specialized Video LLM that builds upon curated 178K synthetic data, encompassing captioning and QA data.
For this model, each video is downsampled to 64 frames, totaling a 64$\times$182 token count.

\paragraph{Qwen2.5-VL-7B}~\cite{qwen2.5-vl} is a frontier MLLM with exceptional multimodal understanding capabilities and a long context window.
By default, we adopt the natural resolution mechanism to avoid distorting the aspect ratio.
We set a maximum frame resolution of 448 and sample 32 frames from a video, yielding at most 32$\times$256 vision tokens per instance.

\subsection{Baseline Details}
\label{sec:baseline_detail}

\paragraph{FastV}~\cite{fastv} is a training-free token pruning strategy designed initially for efficient image understanding.
The method estimates the importance of vision tokens based on the attention score that the token receives from the last query token.

\paragraph{DART}~\cite{dart} first selects pivot tokens based on key or value norms, and then iteratively adds vision tokens less similar to the retained set.
We follow the official implementation by setting both text and vision pivot numbers to 4.

\paragraph{VisionZip}~\cite{visionzip} prunes tokens in the vision encoder instead of the LLM backbone. It first identifies salient tokens based on attention scores, and then merges the remaining ones into contextual tokens based on similarity.
We follow the official implementation by setting the contextual token ratio to 5\%.
The remaining token budget is allocated to salient tokens.

\paragraph{VidCom\textsuperscript{2}}~\cite{vidcom2} designs a two-step pruning framework. 
It first allocates a frame-wise retention ratio based on the similarity between each frame and a global video representation. 
Then, the uniqueness of each token is determined based on its similarity to both the global video representation and the pooled frame representation.

\paragraph{FastVID}~\cite{fastvid} adopts a segment-then-prune framework. Within each segment, salient tokens are first selected frame-wise; the remaining tokens serve as contextual tokens, which are clustered and merged.
We follow the official setting by setting the minimum segment number $c$ to 8 and the similarity threshold $\tau$ to 0.9.  

\paragraph{HoliTom}~\cite{holitom} utilizes a similar segment-then-prune framework. 
The optimal segmentation is estimated by maximizing the overall number of static tokens. 
Following the original setup, we set a static similarity threshold $\tau$ to 0.65 when the retention ratio $\bar{r}$ is 0.1, and to 0.8 for other cases.

\section{Further Discussion and Case Studies}
\label{sec:app_discuss}

\subsection{Attention Patterns}
\label{sec:app_attn}
In the main text, we compare the performance of two variants to extract the attention score, \ie, (1) using the full attention weight matrix and pooling on the query dimension~\cite{holitom,visionzip} (denoted as ``attn''), and (2) using a global query (similar to the \texttt{[CLS]} token) to calculate~\cite{fastvid} (denoted as ``p-attn''). 
We empirically find that ``p-attn'' performs better.
Further discussion of the performance gap is provided here.

\begin{figure*}[p]\centering
\includegraphics[width=0.76\textwidth]{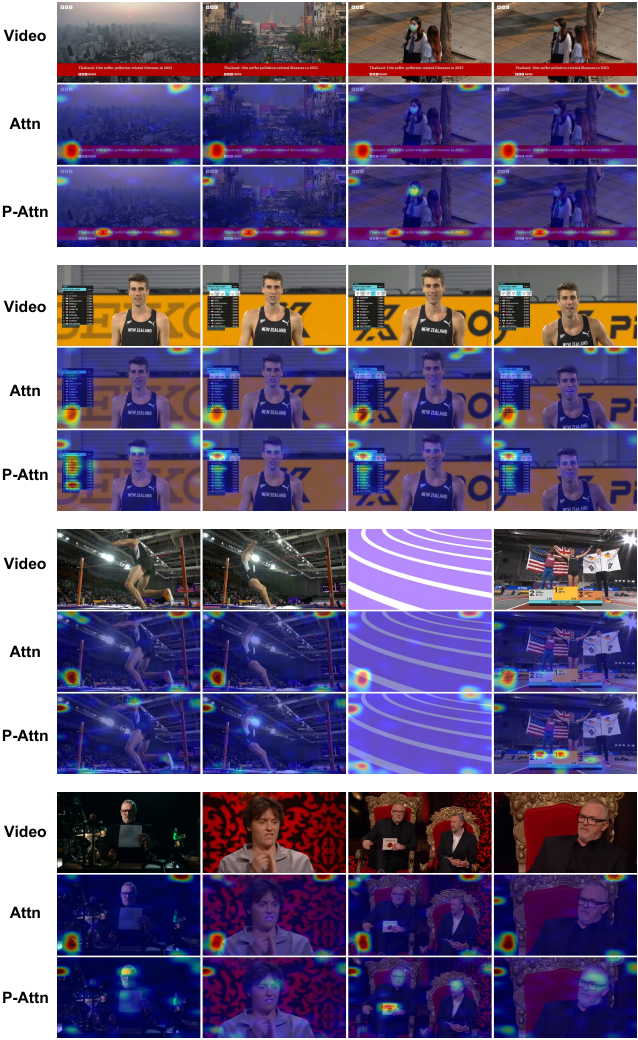}
 \caption{\textbf{Qualitative cases of attention patterns.} We observe that (1) ``attn'' shows more severe attention sinks, failing to attend to salient objects; (2) ``p-attn'' shows a strong tendency to attend to text-related regions. For instance, the subtitle in the first case, and the leaderboard in the second and third cases.}
	\label{fig:attn_case}
\end{figure*}

As shown in~\cref{fig:attn_case}, we observe that: (1) ``attn'' shows more severe vulnerability to the attention sink phenomenon~\cite{darcet2023vision,jiang2025vision}, and fails to capture salient objects in the foreground.
Therefore, Top-$k$ selection based on ``attn'' score can be less effective.
(2) ``p-attn'' shows a strong tendency to attend to text-related regions. This tendency can be advantageous to OCR-related tasks. In fact, Top-$k$ selection with ``p-attn'' scores significantly higher than ``attn'' on OCR problems of Video-MME benchmark (55.4\% \vs 50.4\%).

\subsection{Clustering Results}
\label{sec:app_cluster}
\begin{figure*}[p]\centering
\includegraphics[width=0.78\textwidth]{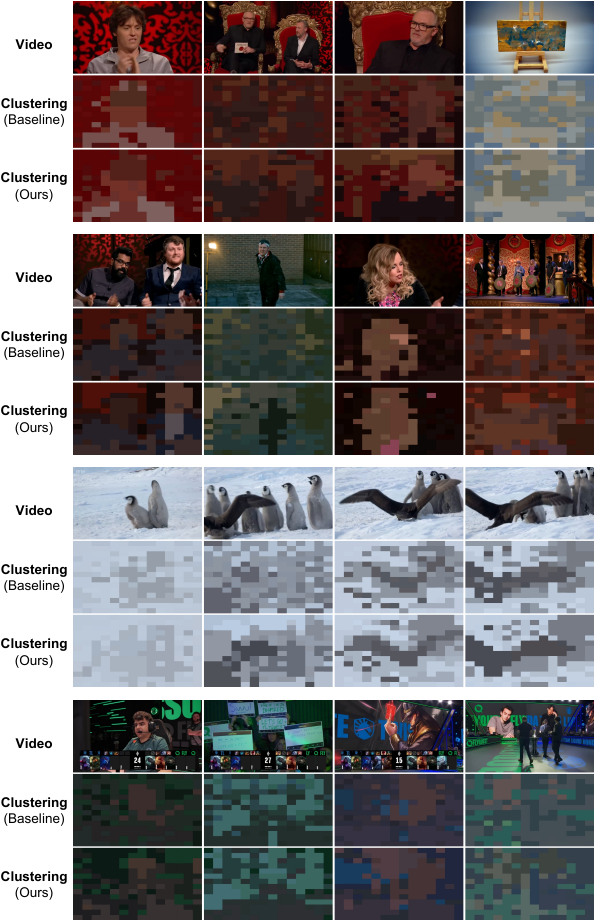}
 \caption{\textbf{Qualitative cases of clustering image features.}
 Our method better preserves the geometric structure of objects and better separates different semantic entities. We note that compressing complex scenes remains a significant challenge (\eg, crowded scenes in the last frame of case 2 or an e-sports event in case 4, which involve intricate or abstract semantics).}
	\label{fig:cluster_case}
\end{figure*}

As shown in~\cref{fig:cluster_case}, our method better preserves object shape and separates different semantic entities.
However, we note that compressing complex open-world scenes remains a significant challenge for efficient video understanding, as it involves perceiving and understanding intricate details and abstract semantics, such as crowded scenes and abstract art.

\begin{figure}[!t]
    \centering
    \setlength{\abovecaptionskip}{2mm}
    \setlength{\belowcaptionskip}{-1mm}
    \includegraphics[width=0.9\columnwidth]{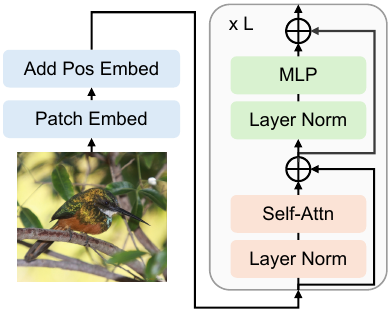} 
    \caption{\textbf{Illustration of SigLIP ViT backbone.} The image is transformed by a patch embedding layer, added with position embedding, then sent into a stack of Transformer blocks, comprising layer normalization, self-attention/MLP, and residual connection.}
    \label{fig:arch_siglip}
\end{figure}

\begin{table}[!t]
\centering
\small
\tablestyle{5pt}{1.15}
\setlength{\aboverulesep}{0.2ex}
\setlength{\belowrulesep}{0.2ex}
\caption{\textbf{Statistics of activation magnitudes} at a hidden state after layer 0. The distribution of sink tokens differs from that of normal tokens, with a larger magnitude scale.}
\label{tab:stat_act}
\begin{tabular}{lcc}
\toprule
\textbf{Metric} & \textbf{Sink Token} & \textbf{Normal Token} \\ \midrule
Top 1           & 212.5         & 5.9             \\
Top 2           & 35.9          & 5.6             \\
Top 3           & 33.5          & 3.7             \\
Top 1\%         & 11.6          & 2.7             \\
Top 10\%        & 1.4           & 0.8             \\
Top 25\%        & 0.7           & 0.5             \\
Mean $\pm$ SD  & 1.0 $\pm$ 6.7 & 0.4 $\pm$ 0.5   \\
Median          & 0.4           & 0.3             \\ \bottomrule
\end{tabular}
% \vspace{-3mm}
\end{table}

\subsection{Investigation of Attention Sink}
\label{sec:app_attn_sink}
As noted previously, the attention sink~\cite{darcet2023vision,jiang2025vision} is a common phenomenon.
Unlike previous findings, we observe that on SigLIP~\cite{siglip} ViT, some attention sink tokens are special in that (1) they are constant in positions across inputs, and (2) their activation magnitudes are generally larger than other tokens (See~\cref{tab:stat_act}), though we do not observe ``massive activations'' (characterized by their scarcity and magnitudes exceeding 100 and approximately 1,000 times larger than the median of their corresponding hidden states~\cite{sun2024massive}).

In the following parts, we discuss the positions and formations of these special tokens and how we mitigate their influence in salient token selection.

\begin{figure*}[!t]
    \centering
    \begin{subfigure}{0.24\linewidth}
        \centering
        \includegraphics[width=\linewidth]{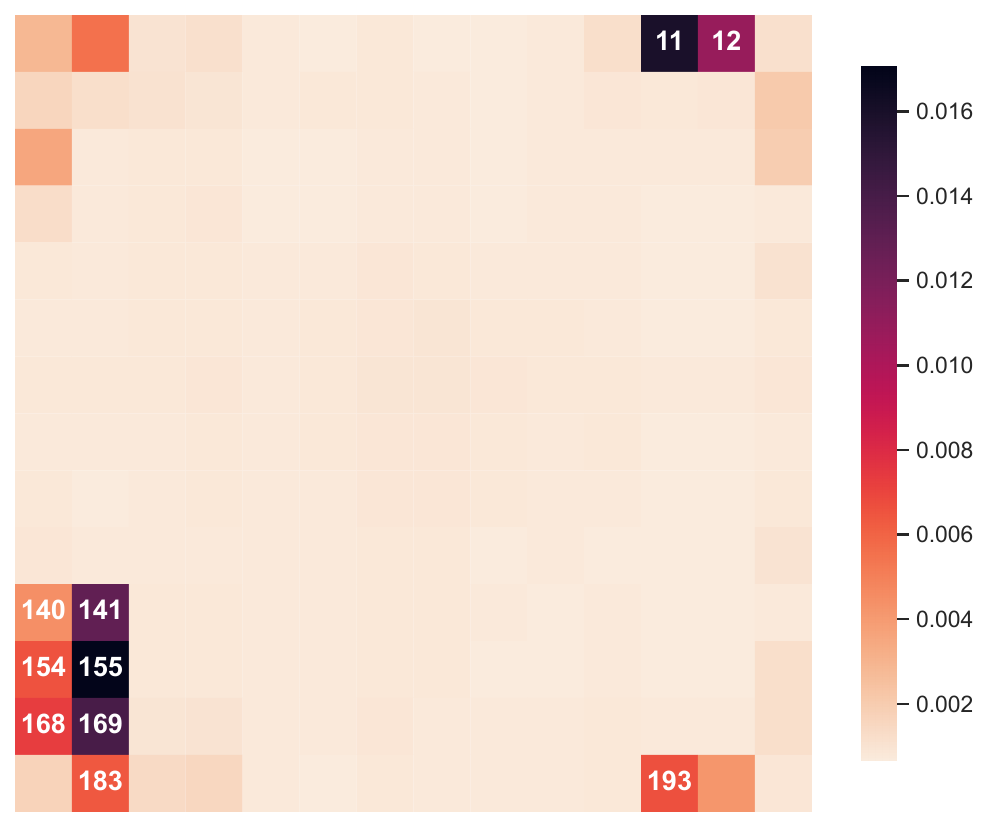}
        \caption{``attn'' spatial distribution in LLaVA-OV-7B.}
    \end{subfigure}\hfill
    \begin{subfigure}{0.24\linewidth}
        \centering
        \includegraphics[width=\linewidth]{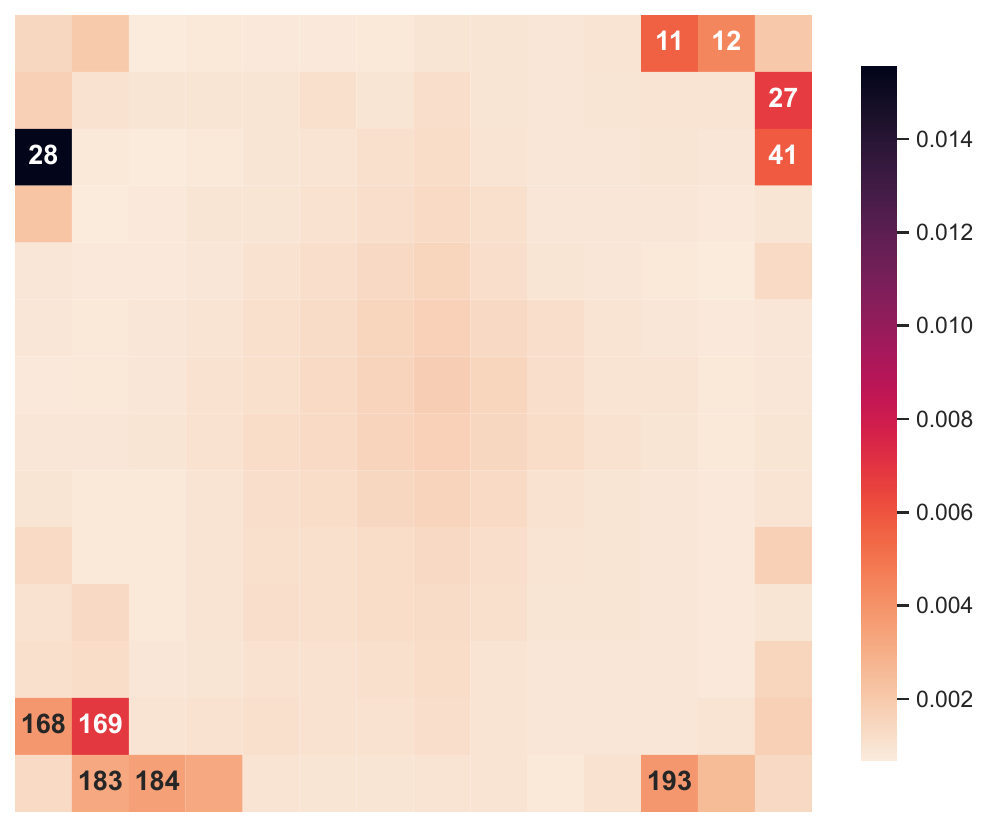}
        \caption{``p-attn'' spatial distribution in LLaVA-OV-7B.}
    \end{subfigure}\hfill
    \begin{subfigure}{0.24\linewidth}
        \centering
        \includegraphics[width=\linewidth]{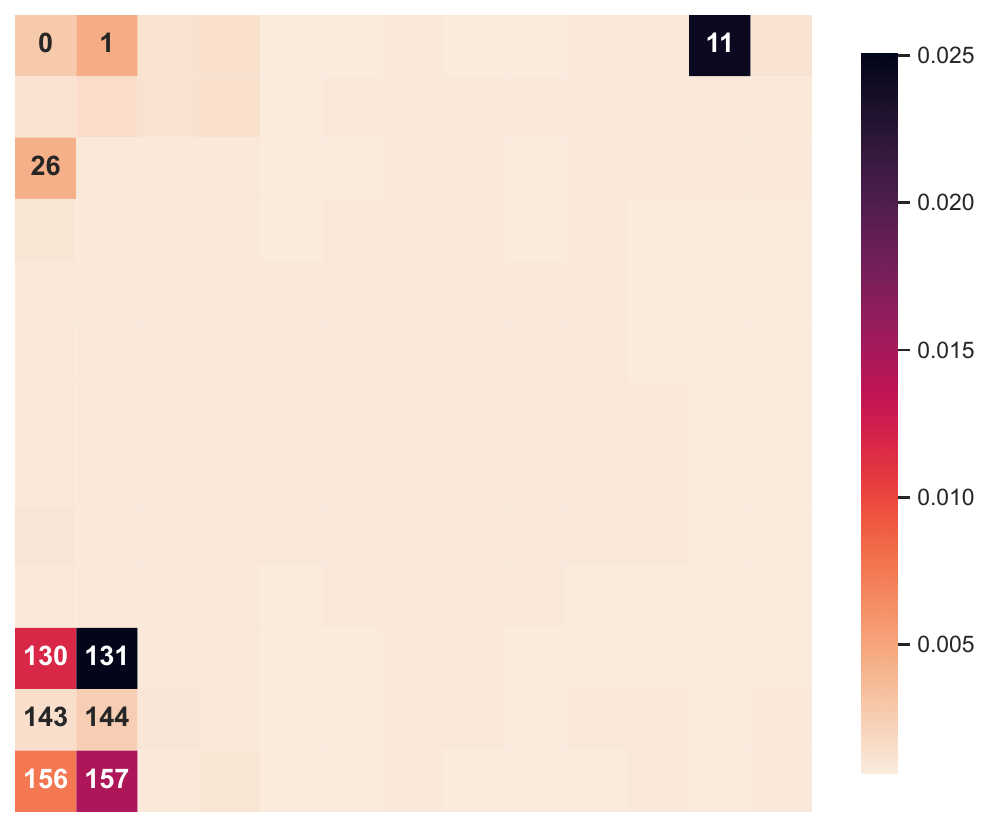}
        \caption{``attn'' spatial distribution in LLaVA-Video-7B.}
    \end{subfigure}\hfill
    \begin{subfigure}{0.24\linewidth}
        \centering
        \includegraphics[width=\linewidth]{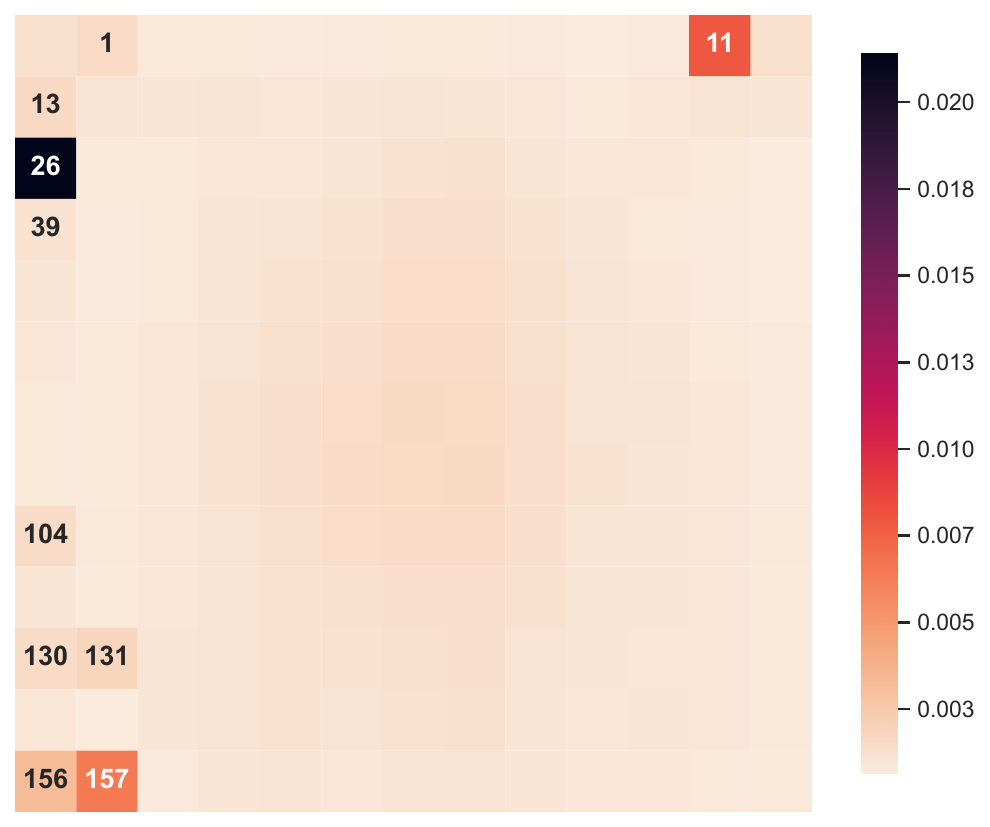}
        \caption{``p-attn'' spatial distribution in LLaVA-Video-7B.}
    \end{subfigure}
    
    \caption{\textbf{Visualization of spatial distributions of attention scores} extracted with SigLIP ViT.
    We observe that (1) sink tokens are distributed around the corners (``attn'' and ``p-attn'' show similar patterns), usually corresponding to backgrounds; and (2) ``p-attn'' shows more robustness to attention sinks.
    The indices of the top-10 highest-scoring tokens are annotated.
    Results are averaged over 100 video samples from ActivityNet~\cite{caba2015activitynet}.}
    \label{fig:attn_sink}
\end{figure*}

\paragraph{Where Are These Tokens?}
We statistically analyze top-ranking attention tokens, and visualize the results in~\cref{fig:attn_sink}.
As shown in the figure, these sink tokens are distributed at the corners of a frame, usually corresponding to backgrounds, and are less relevant to the global semantics.
Moreover, the top-ranking tokens are rather stable. For instance, the rank of \#28 token is $1.28 \pm 0.67$ (mean $\pm$ SD), almost always occupying the highest attention score.
Notably, ``p-attn'' exhibits less vulnerability to attention sinks, which may partially explain the superior performance when used in salient token selection.
To mitigate the impact of attention sinks and retain more relevant tokens, we adopt a simple heuristic, \ie, masking those top-ranking attention sinks that are constant in positions when selecting salient tokens.
We leave explorations of better approaches for future work.

\begin{figure*}[!t]\centering
	\includegraphics[width=0.9\textwidth]{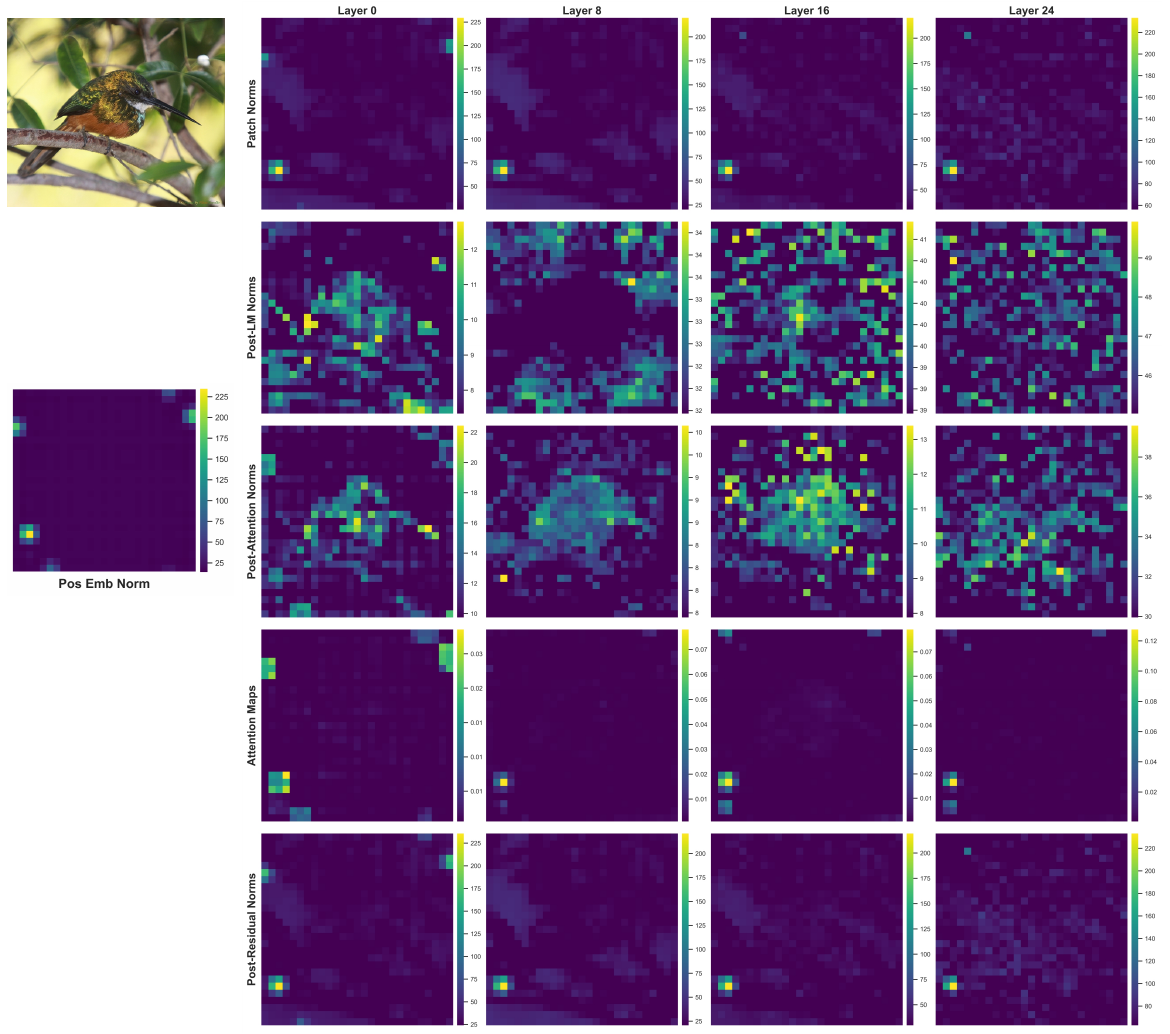}
 \caption{\textbf{Qualitative case of intermediate variables within the attention block.} We observe that (1) the sink tokens are initially induced by adding with the position embedding, where a few outlier tokens exhibit exceptionally high norms; (2) the skewed distribution induced by outliers is passed across layers through residual connection, and temporarily reverted by layer normalization (LM); (3) post-attention norms in shallow layers better preserve low-level geometric than attention score, which is overshadowed by attention sink.}
	\label{fig:layer_case}
\end{figure*}

\paragraph{How Do These Tokens Form?}
We empirically trace the origins of these outlier tokens.
Previous work~\cite{jiang2025vision} attributes these sinks to a sparse set of neurons highly activated by the MLP of a certain layer (\eg, the 6th layer of OpenCLIP ViT-B/16~\cite{jiang2025vision}).
In this work, we find that the sink tokens in SigLIP may originate from a different mechanism. The backbone of the SigLIP ViT is illustrated in~\cref{fig:arch_siglip}.

We visualize the intermediate variables within the attention calculation module, comprising layer normalization, self-attention, and a residual connection.
As shown in~\cref{fig:layer_case}, we observe that (1) the sink tokens are initially induced by adding with the position embedding, where a few outlier tokens exhibit exceptionally high norms; (2) the skewed distribution induced by outliers is passed across layers through residual connection, and temporarily reverted by layer normalization (LM); (3) post-attention norms in shallow layers better preserve low-level geometric than attention score, which is overshadowed by attention sink.

\end{document}